 \documentclass[final,5p,times,twocolumn,number]{elsarticle}

\usepackage{amssymb}

\usepackage{makecell}
\usepackage{multirow}
\journal{Sustainable Computing: Informatics and Systems}

\begin{document}

\begin{frontmatter}

\title{Phoeni6: a Systematic Approach for Evaluating the Energy Consumption of Neural Networks}


 \author[1]{Antônio Oliveira-Filho}
 \affiliation[1]{organization={Departamento de Informática, Universidade do Estado do Rio Grande do Norte},
             city={Mossoró},
             state={Rio Grande do Norte},
             country={Brazil}}
            
\author[2]{Wellington Silva-de-Souza}
\affiliation[2]{organization={Instituto Metrópole Digital, Universidade Federal do Rio Grande do Norte},
             city={Natal},
             state={Rio Grande do Norte},
             country={Brazil}}
 
\author[3]{Carlos Alberto Valderrama Sakuyama}
\affiliation[3]{organization={Department at the Polytechnic, University of Mons},
             city={Mons},
             state={Hainaut},
             country={Belgium}}

\author[4]{Samuel Xavier-de-Souza}
\affiliation[4]{organization={Departamento de Engenharia de Computação e Automação, Universidade Federal do Rio Grande do Norte},
             city={Natal},
             state={Rio Grande do Norte},
             country={Brazil}}

\begin{abstract}
This paper presents Phoeni6, a systematic approach for assessing the energy consumption of neural networks while upholding the principles of fair comparison and reproducibility. Phoeni6 offers a comprehensive solution for managing energy-related data and configurations, ensuring portability, transparency, and coordination during evaluations. The methodology automates energy evaluations through containerized tools, robust database management, and versatile data models.
In the first case study, the energy consumption of AlexNet and MobileNet was compared using raw and resized images. Results showed that MobileNet is up to 6.25\% more energy-efficient for raw images and 2.32\% for resized datasets, while maintaining competitive accuracy levels. In the second study, the impact of image file formats on energy consumption was evaluated. BMP images reduced energy usage by up to 30\% compared to PNG, highlighting the influence of file formats on energy efficiency.
These findings emphasize the importance of Phoeni6 in optimizing energy consumption for diverse neural network applications and establishing sustainable artificial intelligence practices.
\end{abstract}


\end{frontmatter}


\section{Introduction}
\label{sec:intro}

Deep Neural Networks (DNN) are being used with relative success in fields such as computer vision and natural language processing)~\cite{alexnet:2012,paperswithcode:2020}. A growing number of initiatives have been promoting the development of these networks to solve everyday problems, including optimizing resource allocation in energy-constrained environments like wireless sensor networks~\cite{Alqaraghuli_Karan_2024}. There are repositories ~\cite{imagnet:2009,kaggle:2023} with hundreds of networks created and made available in lists ordered by accuracy, which is the primary metric used to assess the quality of each network.

Recently, several works have highlighted the relevance of the energy consumption of neural networks' training and inference phases~\cite{wellington, Estimationofenergy201975,canfederated:2020,towards-power:2019,mitreview:2019,systematicreport:2020,towardssystematic:2020, Energy-policy-NLP:2019,survey-energy2019, Neural-Network-Quantization-2021, Survey-Quantization-2021, HAWQ_Quantization_2019, Survey_Theories_Quantized_2018, State_Neural_Network_Pruning_2020, IKHLASSE20228867,wenninger2022how,jounal-trends-ai-energy-consumption,journal-novel-xgboost}. Their results emphasize that the search for energy efficiency can significantly benefit mobile devices' autonomy and positively affect the financial costs and carbon footprints of large data centers distributed worldwide. These works measure energy to evaluate their technique. 

There is an evident global concern for the energy consumption of software products that affect people’s daily lives---neural networks are becoming one of them. This fact has important implications on the criteria used to choose these products. It is reasonable to say that energy consumption is becoming part of the criteria for selecting neural networks, just as accuracy is. However, unlike the accuracy calculation, which fundamentally depends on the dataset and the procedures used during the training phase, the energy calculation depends on the devices involved. This aspect adds extra challenges to reproducing the results (RR) and making fair comparisons (FC) between different networks~\cite{tinyml2021}.

Evaluating the energy consumption of neural networks while adhering to the principles of Fair Comparison (FC) and Result Reproducibility (RR) presents significant challenges. This process demands the management of substantial volumes of data, including energy metrics and system configurations. Furthermore, the methodology must ensure portability and transparency, making it applicable and accessible across diverse evaluation contexts so that it is possible to answer the following questions:
\begin{itemize}
    \item Which specific monitoring tool or driver for each device manufacturer or model should be used?
    \item How to run the monitoring tool or driver on the host operating system, and what parameters are needed for each case?
    \item Which data format results from the measurements, and how and where should they persist?
\end{itemize}

In recent years, various initiatives have explored methods to measure and optimize the energy consumption of neural networks, ranging from direct measurement techniques to network compression strategies, training parameter adjustments, and hardware modifications. These approaches include automated profiling tools, such as EnergyProfiling~\cite{liu2022energyprofiling} and NeuralPower~\cite{li2022neuralpower}, frameworks for energy efficiency analysis like PowerMeter~\cite{chen2021powermeter}, and systematic studies assessing energy impacts across different architectures and platforms~\cite{wang2020systematic}. While such efforts are fundamental, gaps remain in integrating methodologies that ensure reproducibility and fair comparability of results, especially in diverse scenarios. In this context, Phoeni6 seeks to advance research by proposing an integrated and systematic solution for evaluating and optimizing the energy consumption of neural networks.

To simplify the configuration, monitoring, and management of data related to the evaluation of energy consumption of neural networks while adhering to the FC and RR principles, we propose Phoeni6. This new systematic approach comprises a set of tools to promote portability, transparency, and coordination in the automation of the main stages of the evaluation process.
The contributions to the state of the art stemming from this proposition include:
\begin{itemize}

\item A set of containerized tools to meet operating system portability requirements;

\item A database management system (DBMS) to meet persistence management requirements;

\item A data model to manipulate and persist data in the DBMS;

\item A methodology that automatically guides the evaluation of energy consumption;

\item A comparative case study between two networks to validate Phoeni6.

\end{itemize}

The remaining Sections are organized as follows: Section~\ref{sec:phoeni6} introduces the Phoeni6 approach and the requirements it addresses. Section~\ref{sec:case-study} presents the first case study, focusing on trade-offs between device energy consumption and network accuracy when varying file size and image resolution. Section~\ref{sec:case-study-2} presents the second case study, demonstrating the flexibility and adaptability of Phoeni6 by investigating device energy consumption based on different image file formats.
Section~\ref{sec:related-works} provides an overview of related work, and Section~\ref{sec:conclusion} discusses conclusions and future research directions.


\section{Phoeni6: Methodology and The Set of Tools}
\label{sec:phoeni6}

\begin{figure*}[htbp]
\centering
\caption{Overview of the Phoeni6 system architecture, illustrating the key components and their interactions for energy consumption evaluation. The modular design of Phoeni6 is highlighted, showcasing its scalability potential.}
\includegraphics[width=\textwidth]{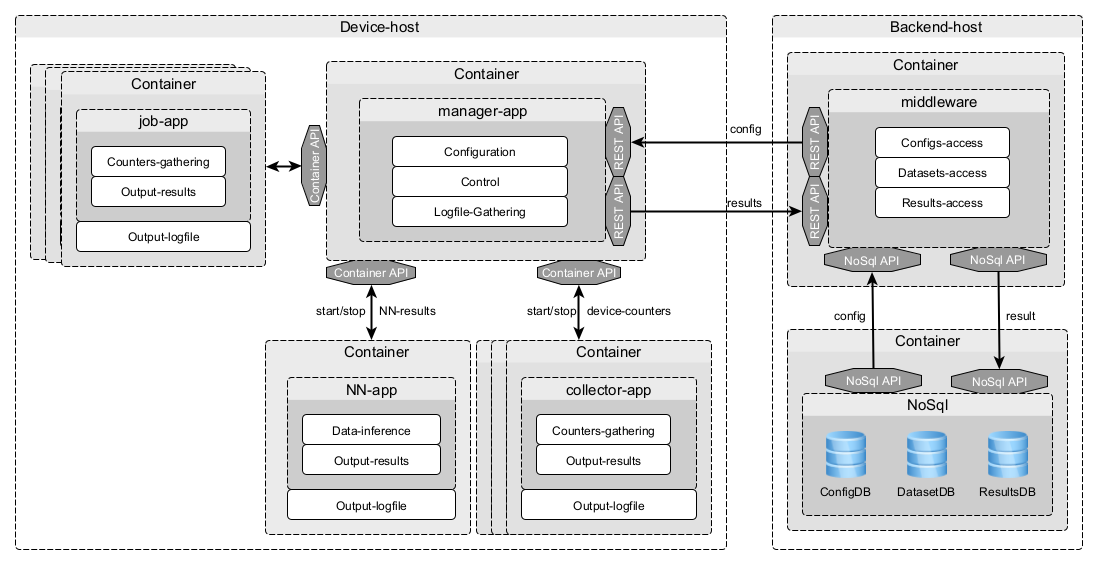}
\label{fig:ecosystem-design}
\end{figure*}

This Section outlines the methodology and components of Phoeni6. Figure~\ref{fig:ecosystem-design} presents the key systems used for assessing the energy consumption of neural networks and is further detailed in the following subsections. The first subsection examines the methodological detail, while the subsequent subsections delve into the tools.

\subsection{Methodology} 
\label{sec:methodology}

\begin{figure*}[th!]
\centering
\caption{Steps of the proposed methodology for energy evaluation, covering all phases from setup to energy calculation.}
\includegraphics[width=\textwidth]{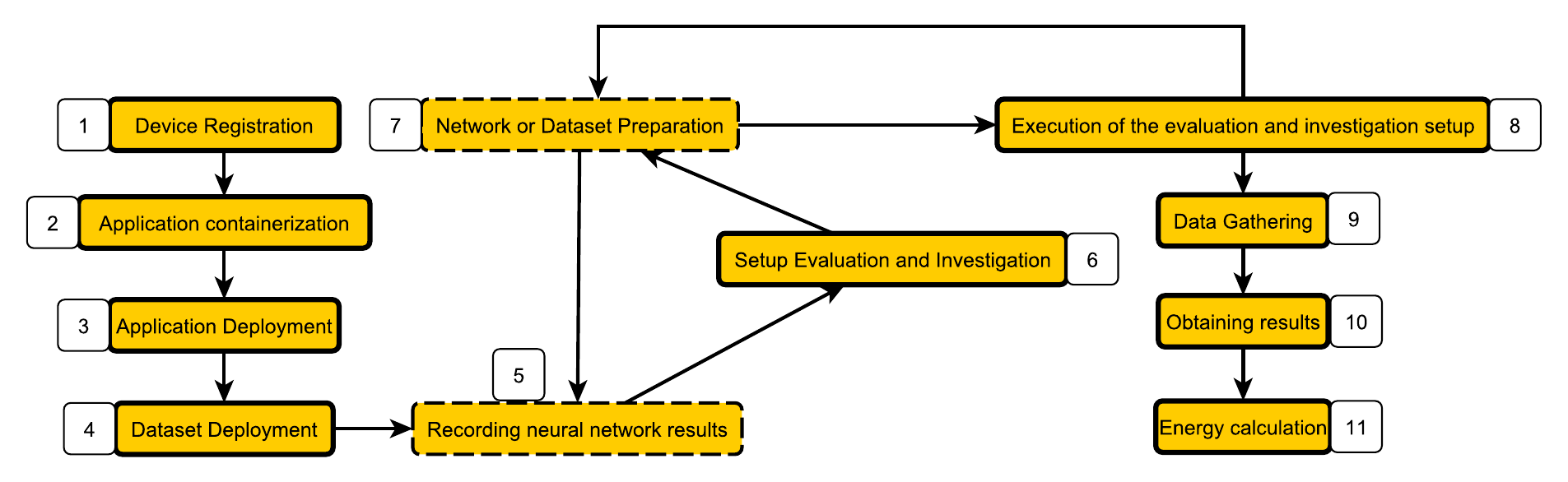}
\label{fig:graph-steps}
\end{figure*}

This methodology is designed to streamline the energy evaluation process of neural networks while ensuring adherence to portability, transparency, and the principles of Fair Comparison (FC) and Result Reproducibility (RR). The proposed approach consists of eleven structured steps, outlined in Figure~\ref{fig:graph-steps}. These steps serve as a guide for users aiming to perform evaluations and leverage the provided tools for process automation. Notably, the methodology remains flexible, allowing subsequent steps to proceed seamlessly even if optional steps are omitted.

\begin{enumerate}
    \item \textbf{Device Registration}: this step refers to the registration of devices in a database so they can be included in future evaluations. The Device entity must be registered according to the model described in Subsubsection-\ref{sec:experimental-data-model}. This way, it will be possible to access information about these devices in an organized and standardized way;

    \item \textbf{Application containerization}: 
    this step consists of the activities required to make an application portable across operating systems and transparent for control by an automated orchestration tool. For example, the collector-app and NN-app, in Figure~\ref{fig:ecosystem-design}, need to go through this procedure. The collector-app is a type of program usually developed by the manufacturer to monitor the performance of one or more devices. For example, nvidia-smi~\cite{nvidia-smi:2020} is a program for some NVidia GPUs. On the other hand, the NN-app is developed by the neural network’s creator or a third party. For everything to work as expected, all dependencies, such as drivers and libraries, must be correctly configured in the container or accessible at runtime.
    
    \item \textbf{Application Deployment}: this step deploys the containerized application on the target execution host for energy consumption assessment. For this, a container manager such as Docker~\cite{docker:2020} or Singularity~\cite{singularity:2020} is required. Additionally, the data identified in the application model described in Subsubsection-\ref{sec:experimental-data-model} are stored in a database for future reference;
    
    \item \textbf{Dataset Deployment}: this step corresponds to deploying a dataset for one or more networks. The dataset must be accessible from the host where the network is or will be deployed, which could eventually be the same host. In addition, configuration data must be recorded in a database according to the dataset entity described in Subsubsection-\ref{sec:experimental-data-model}. Other relevant information, such as the size of each file, data type, and, depending on the type of file, additional information, including image size, number of layers, and others, must be stored in the multivalued attribute. More details are presented in the Subsubsection~\ref{sec:three-databases};

    \item \textbf{Recording neural network results}: In this optional stage, the neural network under evaluation is executed on one of the available datasets. The results of this execution are stored in a database to serve different purposes, including calculating accuracy - a metric commonly used to address trade-offs arising from energy consumption reduction techniques.

Running in either training or inference mode, techniques that modify software or hardware parameters, as well as those that compress the network or dataset to reduce energy consumption, can be evaluated based on the newly calculated accuracy. As Phoeni6 is further utilized, this stage will contribute to creating a results history with the potential to assist in resolving more comprehensive questions, such as those related to multiple models and datasets.

    \item \textbf{Setup Evaluation and Investigation}: this step configures the evaluation and investigation entities described in Subsubsection-\ref{sec:experimental-data-model}. It enables automating the process of obtaining energy consumption results. An evaluation of a network consists of investigations on specific datasets and hardware devices where the neural network runs. This configuration makes the analysis process more agile, allowing accurate and reliable results to be obtained more efficiently;
    
    \item \textbf{Network or Dataset Preparation}: 
    This optional step involves preparation tasks for evaluating or executing each investigation. The tasks include:    
        \begin{itemize}
            \item Modifying the trained NN, such as quantization, pruning, or layer compaction;
            \item Modifying the implanted dataset artifact, such as through compression, or by adding or removing features;
            \item Modifying hardware parameters;
            \item Modifying operating system parameters;
            \item Performing NN training steps;
            \item Modifying the NN during training.
        \end{itemize}
        Each parameter should be stored in a database along with its new value for more precise conclusions about energy consumption.
                
    \item \textbf{Execution of the evaluation and investigation setup}: this step refers to the execution of the configuration saved in step 6.
    
    \item \textbf{Data Gathering}: this step is divided into device and network monitoring. Device monitoring is mandatory and should be transparent and portable so that the program responsible for the reading is containerized. This monitoring corresponds to the reading of the device counters during the network execution and is automated by the process described in Subsection G. The other monitoring is optional. It corresponds to the log data generated during the network execution;
        
    \item \textbf{Obtaining results}: this step consists of submitting the log files generated during the data collection step to the backend host, as shown in Figure 1. As exemplified in Subsection C, the results are directed to a specific route according to their type. This procedure makes the format of the data present in the logs transparent. Finally, all the data are stored in a database, according to the model described in Figure 4, so that it is possible to recover information related to the results from the devices and the program's execution;

    \item \textbf{Energy calculation}: 
        this step uses the data provided in the previous step to perform energy calculations—the Evaluation entity stores the total energy, as shown in Figure 4.    

\end{enumerate}

To ensure scalability across diverse hardware configurations, the Phoeni6 framework was designed with a modular and containerized architecture. This allows individual components, such as the collector-app and NN-app, to be easily adapted for different hardware environments by modifying container configurations. The use of database-driven configurations ensures that the addition of new devices or neural networks can be managed systematically, minimizing manual intervention. For instance, the framework can scale from single-GPU systems to multi-GPU clusters by leveraging parallelized container deployments.

\subsection{Database Architecture}
\label{sec:database-system}

Using a DBMS for data storage and query brings several advantages compared to the ad-hoc mode currently practiced, which uses text files for energy-consumption data storage. Among the main benefits are: 
\begin{itemize}
    \item Increased data integrity and accuracy;  
    \item Improved data retrieval speed and efficiency;  
    \item Easier data manipulation;  
    \item Enhanced data security;  
    \item Higher scalability;  
    \item More accessible data sharing and exchange;  
    \item Improved data backup and recovery capabilities.
\end{itemize}
Among the DBMSs, NoSQLs are more cost-effective in managing large volumes of multivalued data~\cite{mogodb-advantage-nosql:2022}. This type of DBMS is more suitable for supporting energy consumption analysis, as a base data model can be predetermined for the preliminary information, and other kinds of relationships and data aggregations may emerge from its application across different areas. 

As shown in Figure~\ref{fig:ecosystem-design}, our DBMS runs on the same machine as the application that accesses it directly, which ensures more security since the database is not exposed to the internet. Additionally, the system runs in a container to ensure more portability and administrative control over the database. We use tailored databases to create appropriate data models to meet the application’s needs. This approach ensures that the data is stored in an efficient, organized, and secure manner. 

The following subsections present more details on our DBMS implementation.

\subsubsection{Database Structure and Purpose}
\label{sec:three-databases}

Managing large amounts of experiments is one of the main concerns when studying the energy consumption of machine-learning software. Some studies generate billions of records, which need to be kept for future calculations. 
Our proposal includes a set of databases for three primary purposes:
\begin{itemize}
    \item Store the configuration of analysis and experiments;
    \item Store the structure of datasets;
    \item Store the results.
\end{itemize}
Each database contains collections instead of tables in the database system as follows.

\begin{itemize}
    \item \textbf{ConfigDB}: the database used to store data corresponding to the model presented in Figure~\ref{fig:study-model}. Five collections are created, one for each entity of the model: Evaluation, Investigation, Dataset, Containerized\_application, and Device. These collections maintain the necessary configurations to automate energy consumption evaluations outlined in Section~\ref{sec:experimental-data-model}.
\begin{figure}[htbp]
\centering
\caption{Data model for configuring investigations in Phoeni6, illustrating relationships between evaluation components.}
\includegraphics[width=\columnwidth]{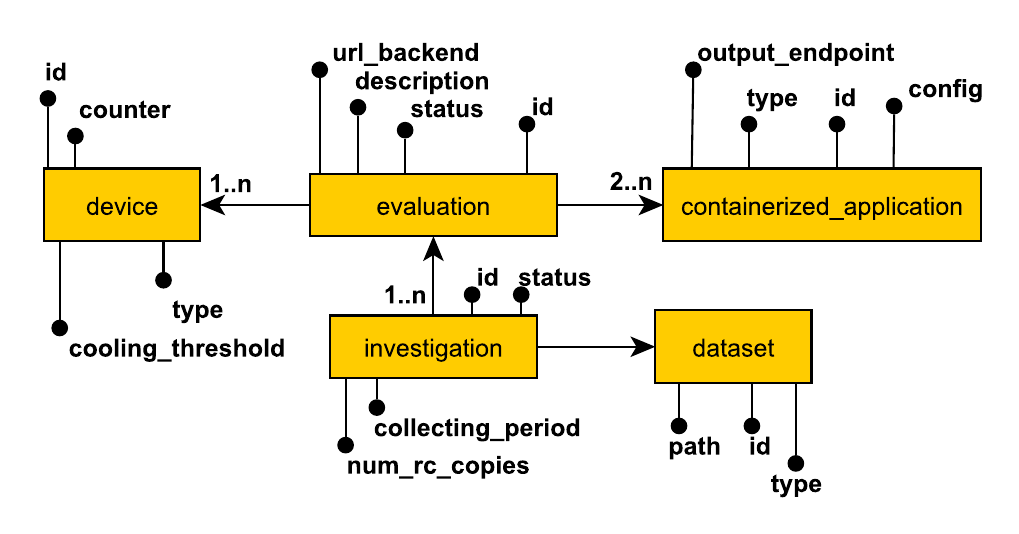}
\label{fig:study-model}
\end{figure}

    \item \textbf{DatasetDB\_[name\_version]}: each database stores the metadata of the datasets used for evaluating the neural networks. For example, ImageNet 2012 is a popular image dataset used for training and testing deep neural networks (DNNs) and can be represented through a database named \texttt{datasetDB\_ImageNet\_2012} that contains collections such as \texttt{dataset\_JPEG}, \texttt{dataset\_PNG}, and so on. These metadata collections can include relevant information about the dataset artifacts, such as the name and size of the files, image dimension, and image classification.
    
    \item \textbf{ResultDB\_[id\_evaluation]}: they are databases used to store the results of evaluations and related configurations. Each database is created using the \texttt{ResultDB\_[id\_evaluation]} model. For example, for Evaluation, \texttt{Id = 1}, the \texttt{ResultDB\_1} database is created. This database automatically creates collections using the \texttt{investigation\_[id\_investigation]} model. That is, for the investigation of \texttt{Id = 1} belonging to Evaluation ID 1, the \texttt{investigation\_1} Collection is created in the \texttt{ResultDB\_1} database. In each Collection, the counters, the respective collected data, and the timestamp associated with each record are stored, in addition to relevant configuration data to characterize each specific investigation.
\end{itemize}

\subsubsection{Structure of Investigation Configuration Data}
\label{sec:experimental-data-model}

Figure~\ref{fig:study-model} gives a general idea of the data model used in the Phoeni6 approach. A brief description of each element and its attributes follows.

\begin{itemize}
    \item \textbf{evaluation}: 
    An evaluation is an integrated set of investigations, devices, and applications, coordinated by the \texttt{manager-app}, which selects the next evaluation to be executed based on the 'pending' status. As described in Figure~\ref{fig:study-model}, an evaluation may consist of two or more applications, with at least one 'Network' and one 'Collector' application being mandatory. Other relevant attributes are detailed in Table~\ref{tab:att_study}.
    
    \begin{table}[htbp]
    \caption{Key attributes for configuring evaluations in Phoeni6, detailing identifiers and descriptions.}
    \centering
    \begin{tabular}{|l|c|l|}
    \hline
         \textbf{attributes} & \textbf{type} & \textbf{description}\\ \hline
         \texttt{id} & string & An unique key \\ \hline
         \texttt{description} & string & The description text \\ \hline
         \texttt{status} & string & A control support \\ \hline
         \texttt{url\_backend} & string & Backend-host URL\\ \hline
    \end{tabular}
    \label{tab:att_study}
    \end{table}
    
    \item \textbf{investigation}:
        investigations are components of Evaluations whose configurations, when executed, produce results that help in understanding the energy behavior of networks for the indicated devices and datasets. The \texttt{status} attribute indicates whether the Investigation is pending, running, or finished, in the same way that the Evaluation status is used. The \texttt{num\_rc\_copies} attribute refers to the number of copies of the artifacts intended for the Investigation, aiming to achieve two objectives: higher accuracy of the results. Table~\ref{tab:att_expemeriment} shows the attributes used to define an Investigation. 

            \begin{table}[htbp]
    \caption{Attributes defining investigations in Phoeni6, including status, number of copies, and collecting periods.}
    \centering
    \begin{tabular}{|l|c|l|}
    \hline
         \textbf{attributes} & \textbf{type} & \textbf{description}\\ \hline
         \texttt{id} & string & A unique investigation id \\ \hline
         \texttt{status} & string & A control support \\ \hline
         num\_rc\_copies & number & \makecell{The number of copies \\ of the  resources}\\ \hline
         \texttt{collecting\_period} & number & The collecting period  \\ \hline
         
    \end{tabular}
    \label{tab:att_expemeriment}
    \end{table}

    \item \textbf{containerized\_application}: an application is an implementation that contributes to the energy consumption evaluation of one or more neural networks. There are three main application groups: those that implement the network itself, those that monitor the devices, and those that perform mediating tasks. In the Phoeni6 approach, all applications need to be containerized. Table~\ref{tab:att_app_nn_model} lists the main attributes that characterize a containerized application. The Type attribute defines which group the application belongs to: Network applications implement the models, Collector applications monitor the device counters, and Job applications perform intermediary tasks for the network under analysis. 
    \begin{table}[htbp]
    \caption{Attributes of applications containerized in Phoeni6, categorized by type and configuration details.}
    \centering
    \begin{tabular}{|l|c|l|}
    \hline
         \textbf{attributes} & \textbf{type} & \textbf{description}\\ \hline
         \texttt{id} & string & \makecell{The name of the  \\          application plus its version} \\ \hline
         \texttt{config} & object & \makecell{A specific container \\  configuration} \\ \hline
         \texttt{output\_endpoint} & string & \makecell{An rest endpoint to send \\ data logging} \\ \hline 
         \texttt{type} & string & network, collector, and job     \\ \hline
    \end{tabular}
    \label{tab:att_app_nn_model}
    \end{table}
    
    \item \textbf{dataset}: it is a deployed Dataset associated with one or more investigations of an evaluation. Table~\ref{tab:att_dataset} shows the attributes that characterize a dataset.
    \begin{table}[htbp]
    \caption{Dataset attributes stored in Phoeni6, including type, path, and version details.}
    \centering
    \begin{tabular}{|l|c|l|}
    \hline
         \textbf{attributes} & \textbf{type} & \textbf{description}\\ \hline         
         \texttt{id} & string & \makecell{The name of the dataset \\ plus its version} \\ \hline
         \texttt{type} & string & Could be Image, Audio, etc. \\ \hline
         \texttt{path} & string & \makecell{The path where the dataset \\ was deployed}  \\ \hline 
    \end{tabular}
    \label{tab:att_dataset}
    \end{table}
    
    \item \textbf{device}: 
    The devices on which the evaluations will be conducted are characterized by the attributes in Table~\ref{tab:att_device}, including the \texttt{cooling\_threshold} attribute. This attribute plays a crucial role in ensuring the accuracy of the results by mitigating the influence of concurrent external processes running on the same devices under study.
    
\end{itemize}    

    \begin{table}[htbp]
    \caption{Key attributes of devices evaluated in Phoeni6, covering identification and energy counters.}
    \centering
    \begin{tabular}{|l|c|l|}
    \hline
         \textbf{attributes} & \textbf{type} & \textbf{description}\\ \hline
         \texttt{id} & string & Unique global identification \\ \hline
         \texttt{counter} & string & \makecell{The power counter\\ of the device}\\ \hline
         \makecell{\texttt{colling\_}\\ \texttt{threshold}} & number & \makecell{A standby device power \\ draw threshold} \\ \hline
         \texttt{type} & string & \makecell{A type of device \\ such as GPU, CPU, etc.} \\ \hline
    \end{tabular}
    \label{tab:att_device}
    \end{table}

\subsubsection{Structure of Investigation Result Data}
\label{sec:result-data-model}

\begin{figure}[htbp]
\centering
\caption{Data model for storing investigation results, detailing attributes for energy consumption evaluation.}
\includegraphics[width=\columnwidth]{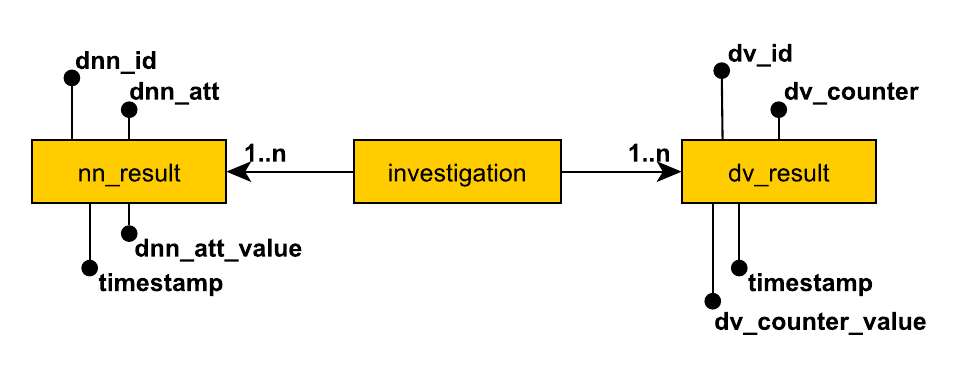}
\label{fig:result-model}
\end{figure}

The definition of models representing the results obtained through investigative processes is just as important as those used to configure the processes themselves. While the latter ensures the easy verification of configurations, the former facilitates the evaluation of the outcomes. In Section~\ref{sec:set-of-tools}, both types of results are addressed in greater detail by the collector-app and NN-app application groups. Each of these applications has specific attributes tailored to its context, as represented in Figure~\ref{fig:result-model}, which will be discussed further.

\begin{itemize}
    \item \textbf{nn\_result}: 
    the \texttt{nn\_result} model can represent two types of results: those derived from the execution of the NN and those associated with the settings defined for the network before the investigative step. The results are expressed as a set of attributes and their respective values, as stated in the model and described in Table~\ref{tab:nn_result}.    
    \begin{table}[htbp]
    \caption{Attributes of neural network results, specifying collected metrics and timestamps.}
    \centering
    \begin{tabular}{|l|c|l|}
    \hline
         \textbf{attributes} & \textbf{type} & \textbf{description}\\ \hline
         \texttt{nn\_id} & string & the NN device \\ \hline
         \texttt{nn\_att} & string & \makecell{a NN attribute collected by \\ the nn\_collector} \\ \hline
         \texttt{nn\_att\_value} & string & \makecell{an NN attribute value}  \\ \hline
         \texttt{timestamp} & timestamp & \makecell{The timestamp \\ of the data result} \\ \hline
    \end{tabular}
    \label{tab:nn_result}
    \end{table}
    \item \textbf{dv\_result}: 
    the \texttt{dv\_esult} model can represent two types of results: those coming from monitoring devices during an investigation and those associated with the pre-investigative settings of the devices. These results are expressed as a set of attributes and their respective values, as shown in the model and defined in Table~\ref{tab:dv_result}. The program responsible for this type of interaction belongs to the collector-app group described in Section~\ref{sec:set-of-tools}.
    
    \begin{table}[htbp]
    \caption{Device result attributes, detailing power counters and associated values.}
    \centering
    \begin{tabular}{|l|c|l|}
    \hline
         \textbf{attributes} & \textbf{type} & \textbf{description}\\ \hline
         \texttt{dv\_id} & string & the \texttt{id} device \\ \hline
         \texttt{dv\_counter} & string & \makecell{the  counter collected} \\ \hline
         \texttt{dv\_counter\_value} & string & \makecell{the counter value}  \\ \hline
         \texttt{timestamp} & timestamp & \makecell{the timestamp \\ of the data result} \\ \hline
    \end{tabular}
    \label{tab:dv_result}
    \end{table}
\end{itemize}

\subsection{Middleware for Data Management}
This system implements functionalities that abstract the complexity of accessing the databases presented in Section~\ref{sec:database-system}, allowing data manipulation to be carried out efficiently, securely, and compliant with the best development practices. Through REST routes, the functionalities are executed by making use of the GET, POST, PUT, PATCH, and DELETE methods, allowing actions such as search, addition, change, removal of data, creation of records, user authentication, and access to specific information to be carried out quickly and easily with the highest level of security~\cite{rest-protocol:2022}. 

The use of the REST approach is demonstrated through the following listed endpoints.

\begin{itemize}
    \item All POST method endpoints accept input data in JSON format and return a response with the status of the operation;
    \item All GET method endpoints return the corresponding resources in JSON format.
\end{itemize}

The list of the main endpoints contains:
\begin{itemize}

\item POST /v1/evaluation: allows creating a new evaluation;

\item GET /v1/evaluation/:id: returns the details of the evaluation associated with the given \texttt{Id};

\item GET /v1/evaluation/:id/energy: returns the details of energy consumption associated with the given \texttt{Id}. The energy calculation is performed by querying the data stored in \texttt{dnn\_result} and \texttt{dv\_result}, as detailed in Subsubsection~\ref{sec:result-data-model}. The algorithm is divided into two parts as described below:

\begin{enumerate}
    \item Data Selection: For each investigation of the same network, select all records from \texttt{dv\_result} whose timestamps fall within the time interval specified in \texttt{dnn\_result}.
    \item Energy Calculation: Apply the formula described in Equation \ref{eq:energy} to obtain the consumed energy. The formula used for energy calculation is expressed as:

\end{enumerate}

\begin{equation}
E = \sum_{k=1}^{n-1} \bar{P}_k \times \frac{(t_{k+1} - t_k)}{1000}
\label{eq:energy}
\end{equation}

where:
\begin{itemize}
    \item $E$ represents the total energy consumed during the evaluation period, measured in joules.
    \item $\bar{P}_k$ is the instantaneous power (in watts) recorded at the instant ${t_k}$. This value is derived from device counters and reflects the instantaneous energy usage during the time interval.
    \item $t_k$ and $t_{k+1}$ are the timestamps (in milliseconds) corresponding to the power measurements $P_k$ and $P_{k+1}$. These timestamps define the duration over which the power is assumed to be constant at $P_k$. These timestamps define the duration over which the power was averaged.
    \item The division by 1000 is a unit conversion factor to ensure the energy result is in joules, as power is measured in watts and time in milliseconds.
\end{itemize}

This formula calculates the cumulative energy consumption by summing up the contributions of energy over discrete time intervals. It assumes that the power consumption remains approximately constant within each interval $[t_k, t_{k+1}]$. Variations in $\bar{P}_k$ over different intervals provide insights into the power consumption dynamics during the process.

\item GET /v1/evaluation/status/:status: returns all evaluations with the given \texttt{status};

\item POST /v1/investigation: allows creating a new investigation;

\item GET /v1/investigation/:id: returns the details of the investigation associated with the given \texttt{Id};
\item GET /v1/investigation/evaluation/:id: returns all investigations associated with the given evaluation \texttt{Id};
\item GET /v1/investigation/evaluation/:id/status/:status: returns all investigations associated with the given evaluation \texttt{Id} and with the given \texttt{status};

\item POST /v1/dataset: allows creating a new dataset;

\item GET /v1/dataset/:id: returns the details of the dataset associated with the given ID;

\item POST /v1/device: allows creating a new device;

\item GET /v1/device/:id: returns the details of the device associated with the given ID;

\item POST /v1/nn/deployed: allows creating a new neural network;

\item GET /v1/nn/:id: returns the details of the neural network associated with the given ID;

\item POST /v1/app: allows creating a new application;

\item GET /v1/app/deployed/:id: returns the details of the application associated with the given ID;

\item POST /v1/app/investigation: allows recording the results of the investigation preparation;

\item POST /v1/app/dataset: allows recording the results of the dataset preparation;

\item POST /v1/app/nn: allows recording the results of the network predictions;

\item POST /v1/app/collector: allows recording the results of the data collecting.

\end{itemize}

\subsection{Toolset for Device-Host Execution}
\label{sec:set-of-tools}

Each subsection in this section outlines the components of the system used for energy evaluation. We analyze the collector-app and containerized systems that collect energy data from the investigated devices. We also discuss the NN-app, the network whose energy consumption is the focus of the Evaluation, and the job-app used for specific tasks related to the Evaluation. Finally, we examine the manager-app, which controls the execution of the other containers in the Evaluation. Figure~\ref{fig:manager-experiment-process} illustrates the general steps and the overall flow of the execution for an energy evaluation, respectively.

\subsubsection{Collector-App: Energy Data Acquisition}
    The collector-app represents a group of systems responsible for collecting data from devices such as CPUs and GPUs for use in energy research and evaluation. The device manufacturers generally develop these systems to be specific to each hardware. For example, the system nvidia-smi~\cite{nvidia-smi:2020} was created by Nvidia to read counters such as "power\_draw" and others. 
    
    To ensure security and portability, the applications in this group are containerized and made available for deployment on the machines of interest. Generally, they share the same workflow: data is collected in a pre-defined period, printed to standard output, and directed to a log file associated with the container during its execution. The data is then stored in various formats, which depend on the specific application, making the current solutions highly coupled to the hardware. The entire process is controlled by the manager-app described in Section ~\ref{sec:manager-app}.

\subsubsection{NN-App: Neural Network Execution}

This application implements a neural network, and three main tasks are performed: 
\begin{itemize}
    \item Access the artifacts of the dataset indicated in the investigation's configuration.
    \item Run the network, either in inference or training mode.
    \item Print the predictions to the standard output, which, as with the collector-app, uses the log file associated with the corresponding container.
\end{itemize}

\subsubsection{Job-App: Task Management System}

    Job-app systems perform intermediate tasks between successive executions of the same evaluation, such as network compression, parameter adjustments for the start of a new training epoch, adjustments to device settings, and so on. All stored results are associated with the evaluation for future analysis, and all job-apps are managed and controlled by the manager-app.
    
\subsubsection{Manager-App: Orchestration and Coordination}
\label{sec:manager-app}

\begin{figure*}
\centering
\caption{This diagram illustrates the coordination of containerized applications during the energy evaluation process. It highlights the steps managed by the Manager-app, including initialization, data collection, warming up devices, neural network execution, and logging results. This workflow ensures adherence to fair comparison (FC) and reproducibility (RR) principles by standardizing the evaluation stages.}
\includegraphics[width=\textwidth]{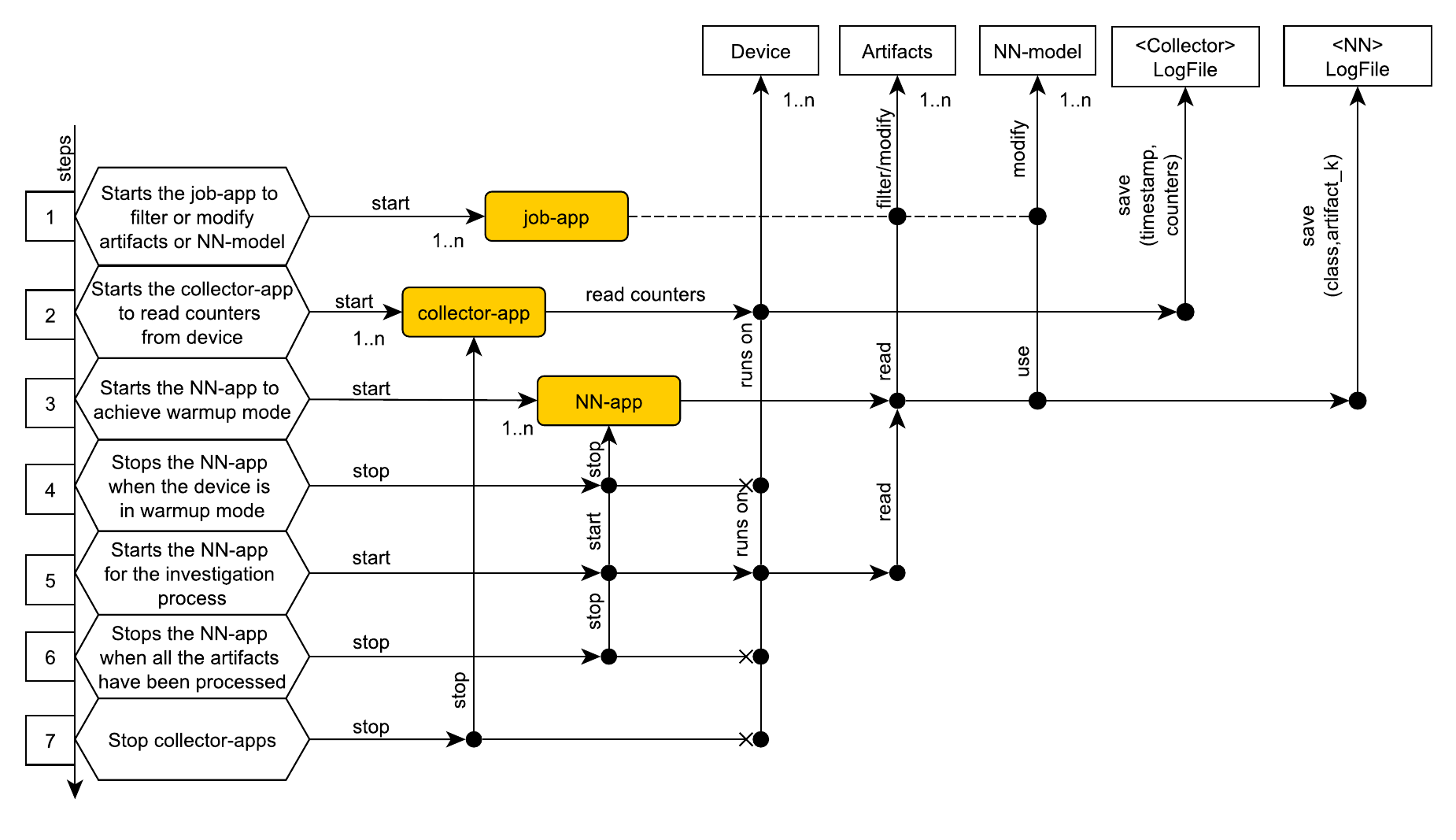}
\label{fig:manager-experiment-process}
\end{figure*}

This application controls the execution of the other containers in the evaluation and implementation configurations. Figure~\ref{fig:manager-experiment-process} shows the general steps and the execution flow of an assessment. Initially, all containers with job-app are executed. Each application, on its initiative, when consulting the values of the assessment's status attributes, decides whether or not to start the programmed tasks. Then, the general process begins, which deals with executing all the experiments aggregated in the current running assessment.   

\begin{itemize}

    \item step 1: all job-apps necessary for the execution of the current evaluation are initiated but remain in standby mode until conditions are suitable for the first one to run;

    \item step 2: the collector-app is started to read the device counters. All collectors related to the ongoing evaluation begin to gather counters, which are logged;
    
    \item steps 3 and 4: help to stabilize the device by warming it up, which allows for fulfilling the FC and RR principles among similar types of devices; this step mitigates the problem of the heating effect, which is different for each device~\cite{cao2020accurate};

    \item step 5: refers to the actual execution of the neural network to conduct the investigations configured for the evaluation.
    
    \item steps 6 and 7: refer to the termination of each investigation.
\end{itemize}

\section{Case Study 1: Energy and Image Processing Trade-offs}
\label{sec:case-study}
Several factors influence the energy consumption of Deep Neural Networks (DNNs) in image classification, particularly image size and resolution. In this section, we employ the Phoeni6 framework to investigate this hypothesis, which also has implications for reducing other computational resources, such as memory and transmission bandwidth. Additionally, we analyze the trade-off between these optimizations and their impact on network accuracy.


\subsection{Models and Dataset Preparation}
\label{sec:choice-models}

The history of the DNN models for the last decade comes across \texttt{AlexNet} and \texttt{ImageNet}. 
\texttt{AlexNet} was the game-change to the neural network area when, in 2012, a team from the University of Toronto improved the TOP 1 Accuracy of image classification by 50\%, the same in a heads or tails game result, to 63\% using an approach of the DNN model~\cite{alexnet:2012}.
Many other models, such as AlexNet, can be checked out at
Papers With Code~\cite{paperswithcode:2020}, including \texttt{MobileNet}, was used to reinforce \texttt{Phoeni6} results beyond comparison.

\begin{figure*}
\centering
\caption{Initial setup steps for the first case study: This diagram illustrates the workflow used in the first case study to evaluate the energy consumption of neural networks when varying image sizes. The process begins with dataset preparation, where the original dataset is filtered and resized to generate additional images for evaluation. Phoeni6 coordinates the classification process, leveraging its modular architecture, including job-apps for resizing and filtering datasets. These steps ensure reproducibility, portability, and scalability of the energy consumption analysis while maintaining consistency across neural network models.}

\includegraphics[width=\textwidth]{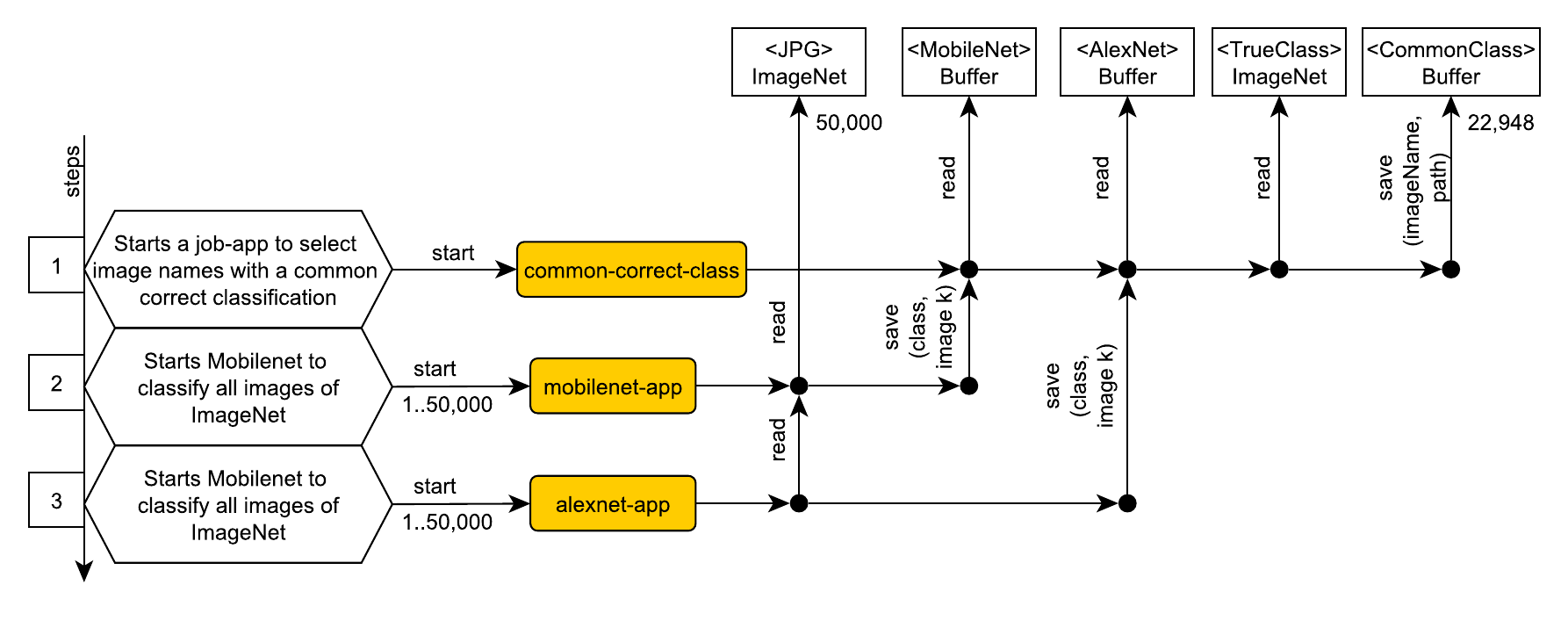}
\label{fig:study1-prepare}
\end{figure*}

\begin{figure*}
\centering
\caption{This figure depicts how the Manager-app orchestrates the tools and processes for the first case study. It emphasizes the modularity and adaptability of Phoeni6, allowing for seamless integration of dataset resizing and energy consumption evaluations, ensuring consistent and reproducible results across different neural networks.}
\includegraphics[width=\textwidth]{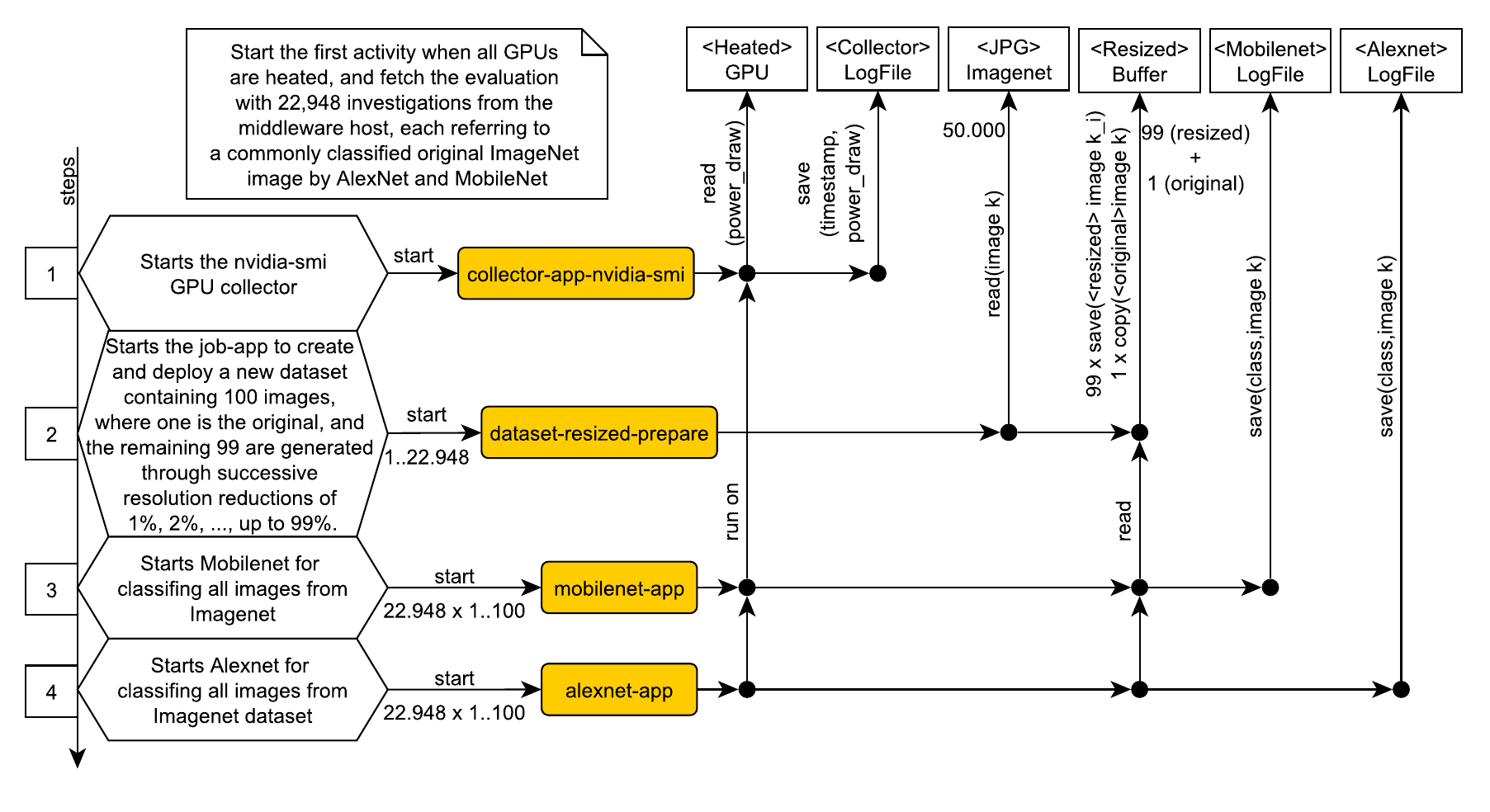}
\label{fig:study1}
\end{figure*}

\subsection{Image Resizing Strategy}
\label{sec:image-approach}
A strategy to answer the raised question is to submit to the networks, in inference mode, a large number of images with different sizes. Although the most widely used datasets offer a good variety of images, they have little variety of sizes. To circumvent this difficulty, we can generate derived images from each original image by reducing them successively. However, this method can take up a lot of disk space. For example, if each image of the Imagenet 2012 test dataset is reduced to 100 new images, in the end, we would have 50,000 x 100 files in total.  



In scenarios like the one presented, the Phoeni6 approach can be essential for developing solutions that mitigate the occurrence of some of these problems, especially when the extra steps can be performed in an interleaved manner during the execution of the main process, which in this case would have an incremental behavior.

The complete solution involves two stages, which may occur at different times. The first stage is described in Figure~\ref{fig:study1-prepare}, and the second is detailed in Figure~\ref{fig:study1}. In the first stage, Phoeni6 is used to classify the images through network execution, with the results being analyzed and segmented by the job-app called common-correct-class, which selects the names and paths of images whose classifications match in both networks. In the second stage, Phoeni6 is employed to perform successive image dimension reductions using the job-app dataset-resized-prepare, with the energy consumption results being collected and stored.


It is important to highlight that the app-jobs mentioned were added as intermediaries to address the unique challenges of this study. If they were not required, the application would run directly on the original dataset without the need for intermediate stages. This demonstrates the flexibility of Phoeni6, as it can adapt to the specific issues being addressed.

The manage-app coordinates all these applications during an evaluation process according to the previously defined execution order. The instances of both applications are represented in Section~\ref{sec:evaluation-setup}.

\subsection{Experimental Setup}

The experiments are developed based on the methodology of Section~\ref{sec:methodology}. The experimental setup is divided into the following Sections.

\subsubsection{Hardware and Software Configuration}

The studies were performed on a server (250 W
max) connected via PCIe 3.0 interface. 
Each server also had
One Intel(R) Core(TM) i7-6700 CPU @ 3.40GHz four cores and eight threads, 8192 KB cache size,
the total memory size was 16 GB.
The server also had two 32 GB Nvidia TITAN X GPUs, each one with the properties values presented in Table~\ref{tab:nvidia-sasha}~\cite{nvidia-smi-properties:2022}:
\begin{table}[htbp]
\caption{Nvidia GPU specifications, including power limits, driver version, and operational settings, essential for reproducibility and energy analysis.}
\centering
\begin{tabular}{ |c|c| } 
\hline
Property & Value \\ \hline
\texttt{power.limit}  & \textbf{189.00 W} \\ \hline
\texttt{power.max\_limit} & \textbf{227.00 W} \\ \hline
\texttt{power.min\_limit} & \textbf{150.00 W} \\ \hline
\texttt{driver\_version} & \textbf{440.82} \\ \hline
\texttt{display\_active} & \textbf{Disabled} \\ \hline
\texttt{persistence\_mode} & \textbf{Disabled} \\ \hline
\texttt{accounting.mode} & \textbf{Disabled} \\ \hline
\texttt{gom.current} & \textbf{\shortstack{Low Double \\ Precision}} \\ \hline
\texttt{compute\_mode} & \textbf{Default} \\ \hline
\texttt{clocks.applications.graphics} & \textbf{705 MHz} \\ \hline
\texttt{clocks.applications.memory} & \textbf{3505 MHz} \\ \hline
\end{tabular}
\label{tab:nvidia-sasha}
\end{table}

These properties are stored on the database system during \textbf{stage 4.1} described in Section~\ref{sec:methodology}. 

\vspace{4mm} 

The main software used in experiments are listed below: 

\begin{table}[htbp]
\caption{Hardware and software setup details used in the experiments, including server specifications and main tools.}
\centering
\begin{tabular}{ |c|c| } 
\hline
Property & value \\ \hline
setup type  & enterprise grade server \\ \hline
operating system & Ubuntu 18.04.4 LTS \\ \hline
OS kernel & Linux Kernel 4.15.0-69 \\ \hline
CPU set & 1x Intel Core i7 6700 \\ \hline
cache & 1 x 8 MB\\ \hline
RAM & 4 x 4 GB \\ \hline
RAM type & DDR4 \\ \hline
GPU set & \shortstack{2x Geforce GTX \\ TITAN z} \\ \hline
\end{tabular}
\label{table:softwarehardwaresetup}
\end{table}

\begin{itemize}
    \item \textbf{Operational System}: Ubuntu 18.04.4 LTS (GNU/Linux 4.15.0-69-generic x86\_64)
    \item \textbf{Models Framework}: Tensorflow tensorflow/tensorflow 1.12.0-gpu-py3;  
    \item \textbf{\texttt{Manager-app} Language}: Python 3.8.0;
    \item \textbf{Middleware Language}: Node.js version 14.16.0. Node.js is a JavaScript runtime built on Chrome's V8 JavaScript engine; 
    \item \textbf{Container Server}: Docker 19.03.10, build 9424aeaee9~\cite{docker:2020}; 
    \item \textbf{Database System}: ArangoDB 3.6.2~\cite{arangodb:2021}.
\end{itemize}

\subsection{Evaluation Setup}
\label{sec:evaluation-setup}

Table~\ref{tab:evaluation} presents the general evaluation settings, including the goal of answering questions about the impact of image size on energy consumption. Table~\ref{tab:device} describes the device used, while Table~\ref{tab:dataset} details the dataset configuration. Finally, Table~\ref{tab:applications} lists the applications employed, categorizing them by type and function.

\begin{table*}[ht]
\centering
\caption{Evaluation Setup}
\begin{tabular}{|l|l|}
\hline
\textbf{Attribute}   & \textbf{Value}                                               \\ \hline
ID                   & 460259218                                                   \\ \hline
Description          & \begin{tabular}[c]{@{}l@{}}Answer questions:\\ - Does image size (height x width)\\ affect DNNs' power consumption?\\ - Which DNN consumes less energy: AlexNet or MobileNet?\end{tabular} \\ \hline
Backend URL          & http://lapps.imd.ufrn.br/phoeni6         \\ \hline
\end{tabular}
\label{tab:evaluation}
\end{table*}

\begin{table*}[ht]
\centering
\caption{Device Configuration}
\begin{tabular}{|l|l|}
\hline
\textbf{Attribute}        & \textbf{Value}                  \\ \hline
Device ID                 & GPU-b24b87b6-fce8-397d-5c35-2fb3bb4fdce2 \\ \hline
Type                      & GPU                             \\ \hline
Cooling Threshold         & 32W                             \\ \hline
Counters Monitored        & Power Draw                     \\ \hline
\end{tabular}
\label{tab:device}
\end{table*}

\begin{table*}[ht]
\centering
\caption{Dataset Configuration}
\begin{tabular}{|l|l|}
\hline
\textbf{Attribute}       & \textbf{Value}                                                  \\ \hline
Dataset ID               & imagenet-2012-[i] \% for i in 45986 \%                          \\ \hline
Type                     & Image File                                                    \\ \hline
Path                     & /phoeni6/dataset/imagenet-2012-[i]                             \\ \hline
Generated Datasets       & 45,986 new datasets                                           \\ \hline
\end{tabular}
\label{tab:dataset}
\end{table*}

\begin{table*}[ht]
\centering
\caption{Application Configuration}
\begin{tabular}{|l|l|l|}
\hline
\textbf{Application} & \textbf{ID}                   & \textbf{Type}     \\ \hline
MobileNet App        & mobilenet-2012               & DNN              \\ \hline
AlexNet App          & alexnet-2012                 & DNN              \\ \hline
Collector App        & nvidia-smi-container         & Collector        \\ \hline
Investigation App    & investigation-prepare-2022   & Job-App          \\ \hline
Dataset App          & dataset-prepare-2022         & Job-App          \\ \hline
\end{tabular}
\label{tab:applications}
\end{table*}

\subsection{Results and Analysis}

The results are based on the application of the Phoeni6 approach, which can be summarized as follows: 
\begin{enumerate}
    \item Through dataset-by-filter-prepare, 22,948 images were obtained, forming the base dataset used in the next step.  
    \item For each image, through dataset-resize-prepare, a new dataset of 100 images was obtained in each iteration, one original and 99 others obtained from successive reductions in their dimensions, which will be used in the next step;  
    \item Each image is copied 100 times to proceed to the next step;  
    \item The network runs in inference mode for the 100 copies. 
It is assumed by convention that the first 50 images serve to warm up the device, and the other 50 are used for calculating the energy consumption;  
\end{enumerate}
At the end of the entire evaluation process of each network, 22,948 (filtered dataset) x 100 (resized dataset) x 100 (copies of each image) were run, resulting in 22,948 x 100 investigations.

\subsubsection{Statistical Analysis Methodology}

The analysis employed the \texttt{RobustScaler} technique to standardize the results of both AlexNet and MobileNet, ensuring variable homogeneity and improving the robustness of the linear regression results. Unlike the Z-score method, which is sensitive to outliers by centering and scaling based on the mean and standard deviation, \texttt{RobustScaler} uses the median and interquartile range (IQR). This approach is particularly relevant when dealing with data that may contain outliers or when the distribution is skewed, as it reduces the influence of extreme values.

Tables~\ref{tab:alexnet-lr-p-value} and~\ref{tab:mobilenet-lr-p-value} illustrate that image size is a key factor influencing the energy consumption of both neural networks, with a notably greater impact observed for AlexNet compared to MobileNet. These findings are supported by the Ordinary Least Squares (OLS) regression analysis, which models the linear relationship between image size and energy consumption. OLS regression helps quantify the strength and direction of this relationship by minimizing the sum of the squared differences between observed and predicted values. The results confirm the hypothesis that more complex tasks, involving larger image sizes, necessitate increased computational processing, thereby leading to higher energy consumption. The OLS regression results, including confidence intervals and regression coefficients, provide a clear statistical basis for understanding how image size affects energy efficiency across different neural network architectures.

\begin{table*}[ht]
\centering
\caption{OLS regression results for AlexNet, analyzing the relationship between image size and energy consumption.}
\begin{tabular}{|l|c|c|c|c|cc|}
\hline
Variable & Coefficient & Standard Error & t-statistic & $P>|t|$ & [0.025 & 0.975] \\
\hline
const              & 0.0009       & 0.001                                                                  & 0.668  & 0.504  & -0.002        & 0.003 \\
image size         & 0.9634       & 0.002                                                                & 564.676  & 0.000  & 0.960         & 0.967 \\
\hline
\end{tabular}
\label{tab:alexnet-lr-p-value}
\end{table*}

\begin{table*}[ht]
\centering
\caption{OLS regression results for MobileNet, evaluating the impact of image size on energy consumption.}
\begin{tabular}{|l|c|c|c|c|cc|}
\hline
Variable & Coefficient & Standard Error & t-statistic & $P >|t|$ & [0.025 & 0.975] \\
\hline
const & 0.0010 & 0.004 & 0.246 & 0.806 & -0.007 & 0.009 \\
image size & 0.6677 & 0.004 & 151.345 & 0.000 & 0.659 & 0.676 \\
\hline
\end{tabular}
\label{tab:mobilenet-lr-p-value}
\end{table*}

The linear relationship between image size and energy consumption highlights the direct impact of input complexity on computational demands. Larger images increase memory usage and processing requirements, leading to higher power draw during inference. This trend is more pronounced in AlexNet, suggesting that its architecture is less optimized for energy efficiency compared to MobileNet. These findings emphasize the importance of considering input size as a critical factor when optimizing neural networks for energy efficiency.

\subsubsection{Energy Comparison: AlexNet vs MobileNet}

\begin{figure}[htbp]
\centering
\caption{Relationship between energy consumption and file size for raw images, with a linear regression model for trend analysis.}
\includegraphics[width=\columnwidth]{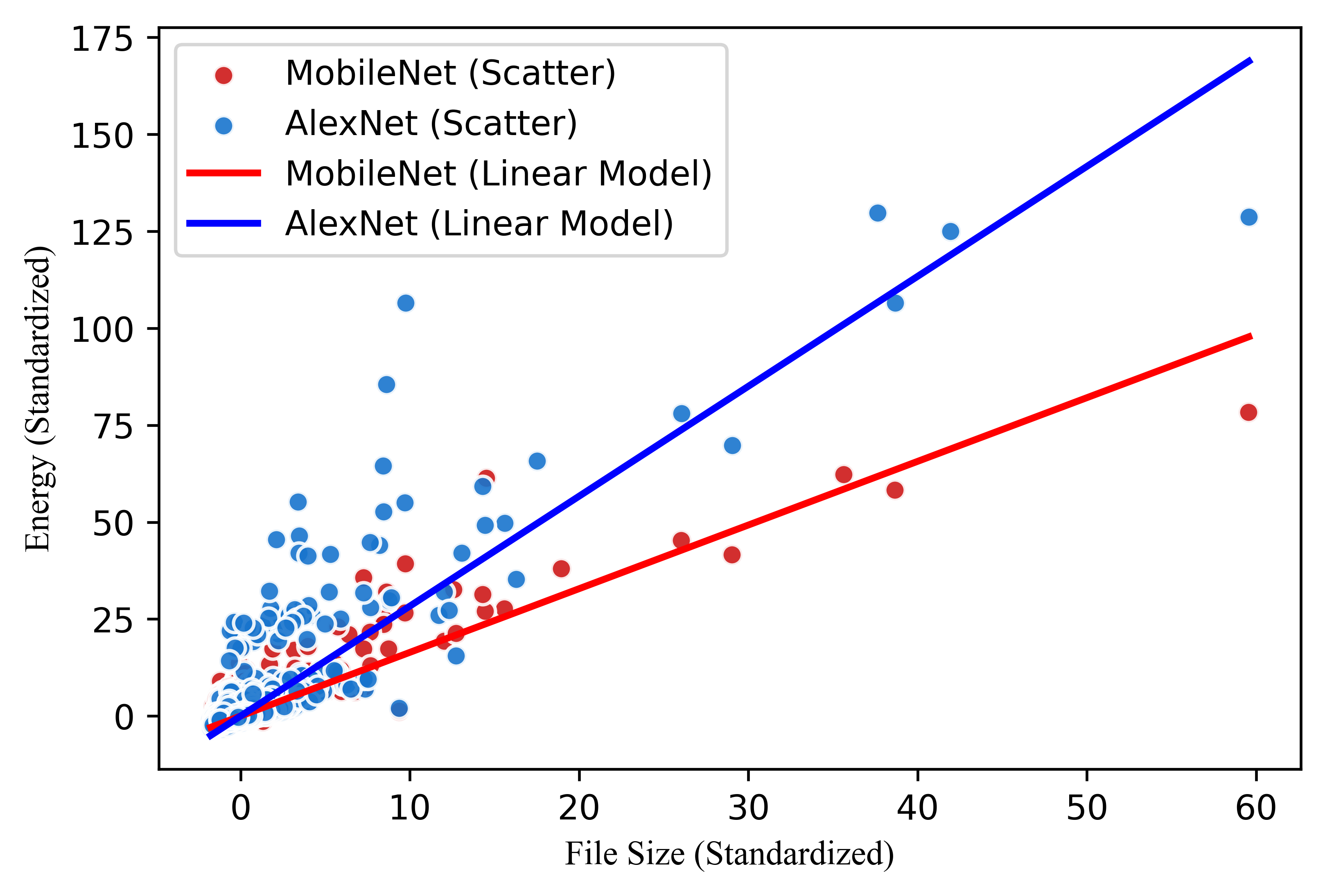}
\label{fig:alexNet_MobileNet_scatter_energy_file_size_raw}
\end{figure}

\begin{figure}[htbp]
\centering
\caption{Scatter plot of energy consumption for AlexNet and MobileNet with raw image data, showcasing regression trends.}
\includegraphics[width=\columnwidth]{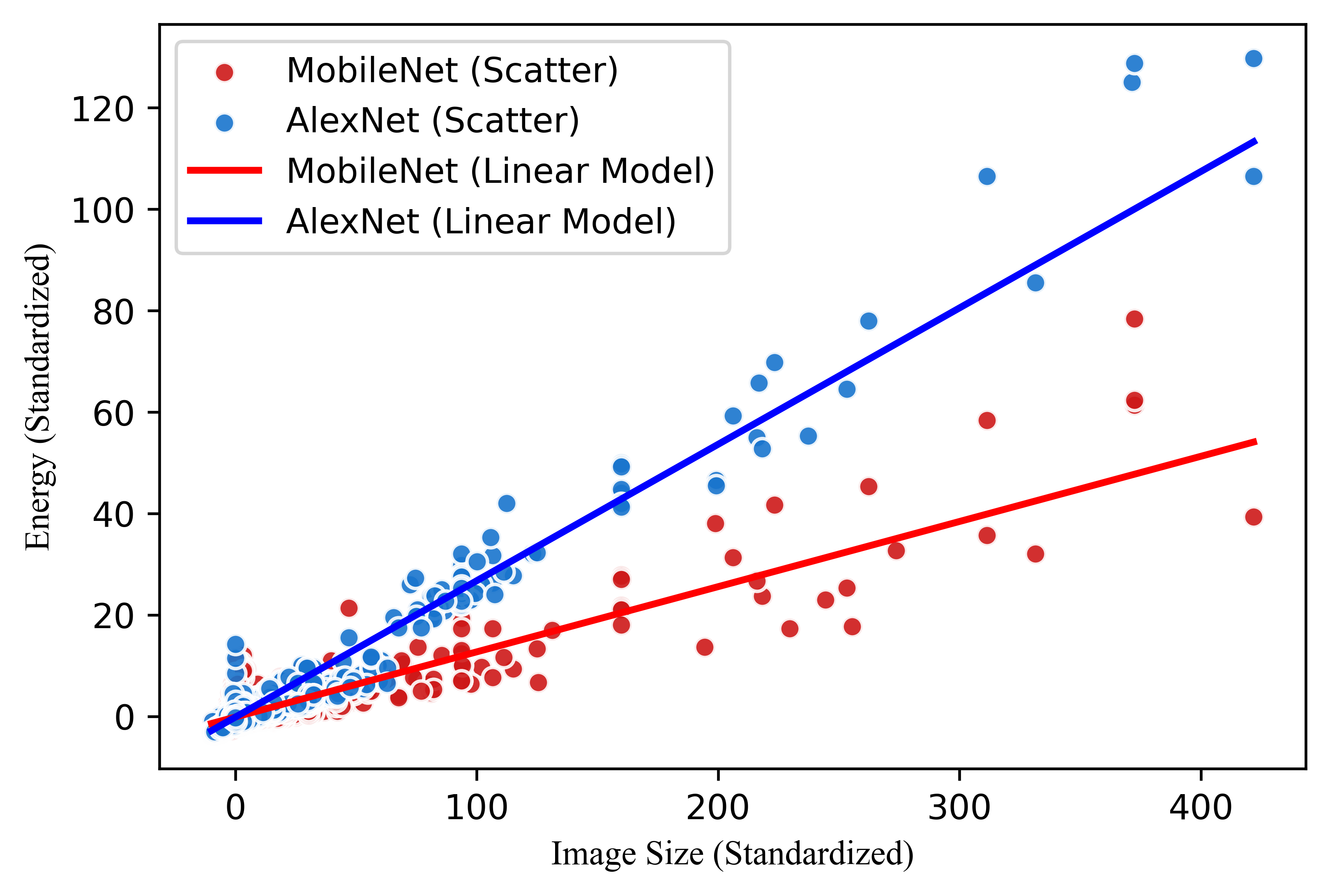}
\label{fig:alexNet_MobileNet_scatter_energy_image_size_raw}
\end{figure}

\begin{figure}[htbp]
\centering
\caption{Analysis of energy consumption as a function of file size for resized images.}
\includegraphics[width=\columnwidth]{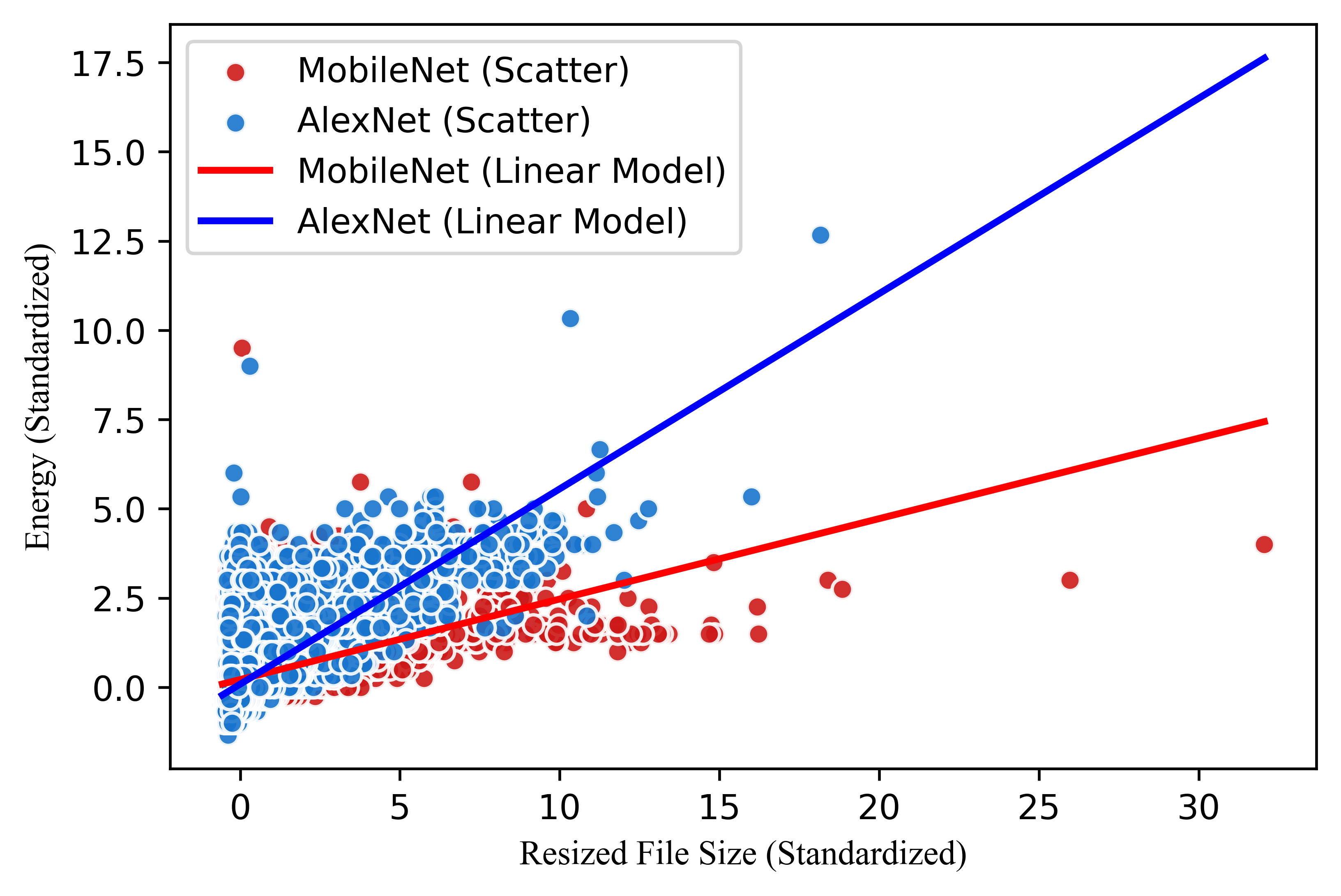}
\label{fig:alexNet_mobileNet_scatter_energy_file_size_resized}
\end{figure}

\begin{figure}[htbp]
\centering
\caption{Relationship between energy consumption and image size for resized datasets, showcasing linear trends.}
\includegraphics[width=\columnwidth]{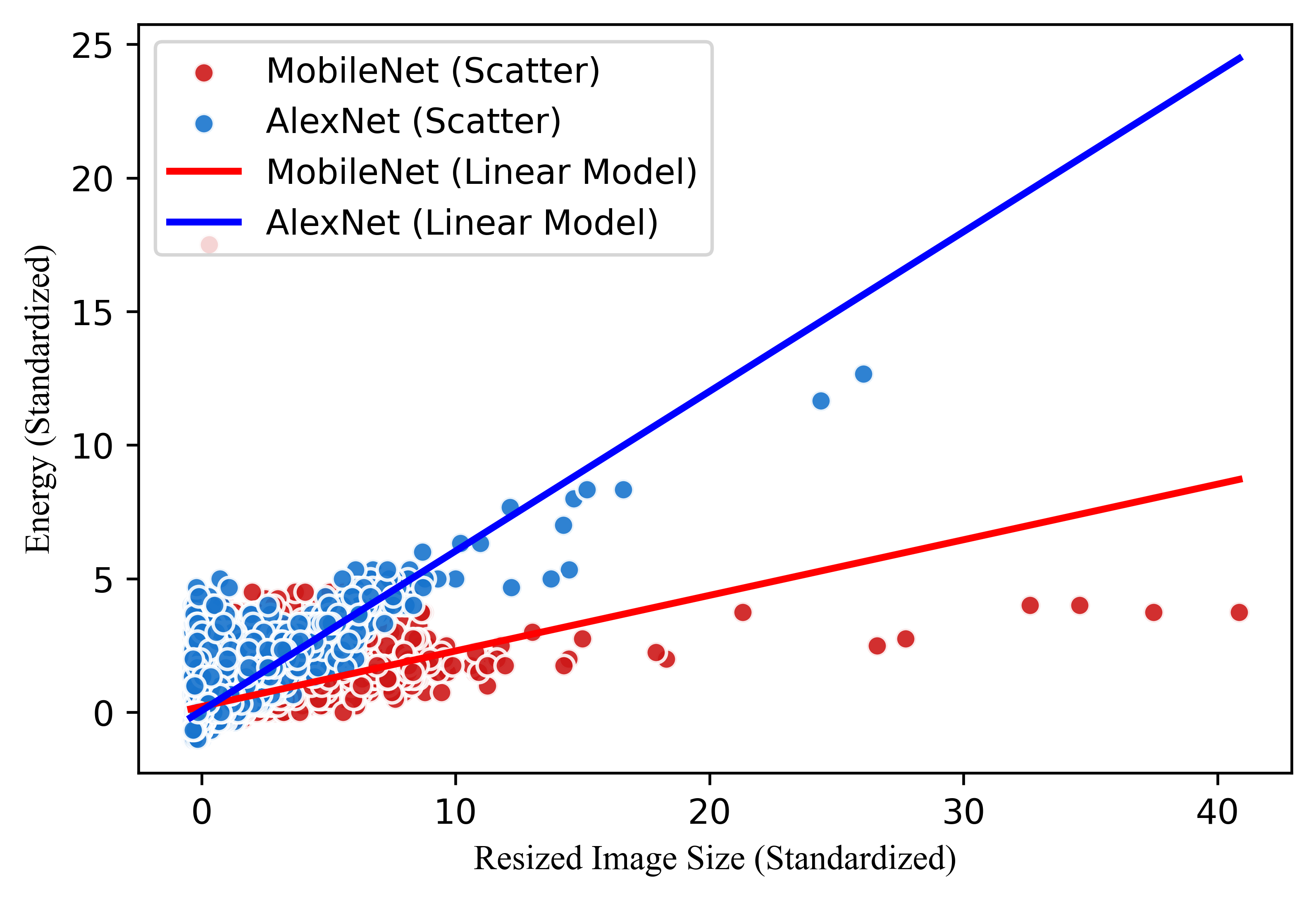}
\label{fig:alexNet_mobileNet_scatter_energy_image_size_resized}
\end{figure}

\begin{figure}[htbp]
\centering
\caption{Energy consumption as a function of file size for the extended dataset (raw + resized images).}
\includegraphics[width=\columnwidth]{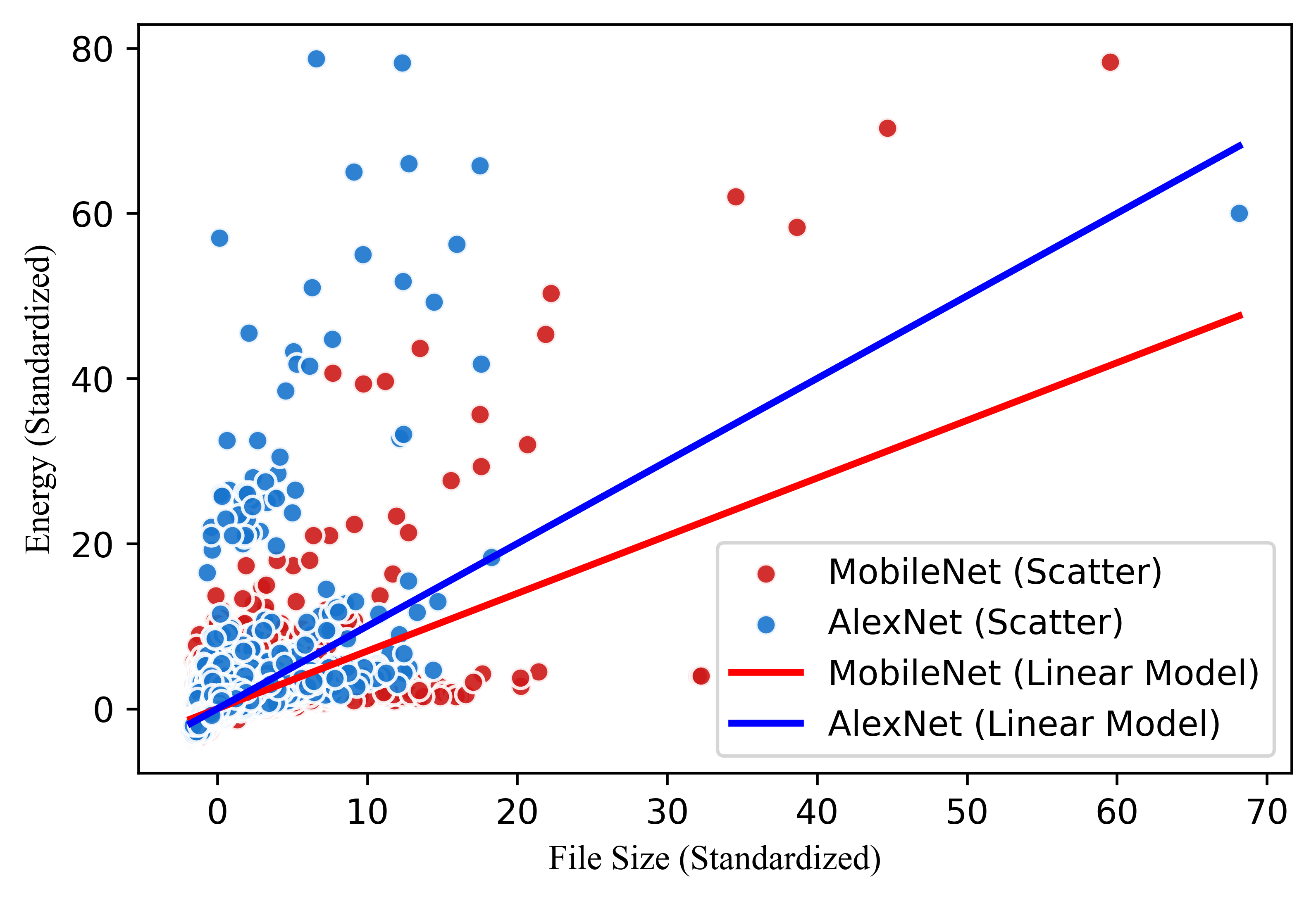}
\label{fig:alexNet_mobileNet_scatter_energy_file_size_extended}
\end{figure}

\begin{figure}[htbp]
\centering
\caption{Energy usage trends for extended datasets, with linear regression analysis of image size impact.}
\includegraphics[width=\columnwidth]{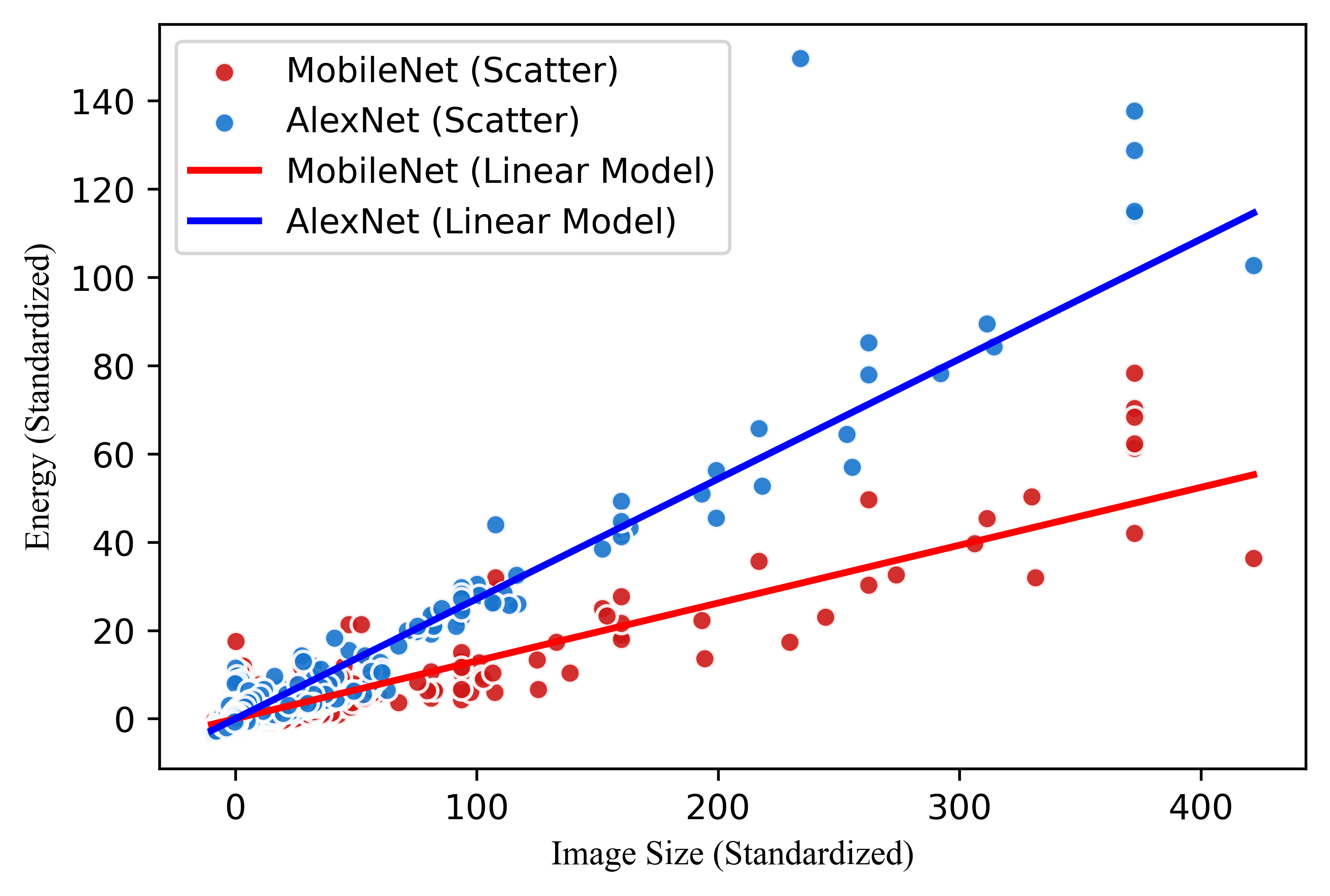}
\label{fig:alexNet_mobileNet_scatter_energy_image_size_extended}
\end{figure}

Based on the provided information, the study results indicate that MobileNet outperforms AlexNet regarding energy consumption in all arrangements. The data is expressed in Figures~\ref{fig:alexNet_MobileNet_scatter_energy_file_size_raw},~\ref{fig:alexNet_MobileNet_scatter_energy_image_size_raw},~\ref{fig:alexNet_mobileNet_scatter_energy_file_size_extended},~\ref{fig:alexNet_mobileNet_scatter_energy_image_size_extended},~\ref{fig:alexNet_mobileNet_scatter_energy_file_size_resized} and~ \ref{fig:alexNet_mobileNet_scatter_energy_image_size_resized}, which display the dispersion of data related to energy consumption. Each graph contains the linear regression corresponding to the respective data scatter, allowing for a clearer visualization of the relationship between image size and energy consumption. These regressions further highlight the more significant impact of image size on AlexNet compared to MobileNet.

In the first arrangement described in Figures ~\ref{fig:alexNet_MobileNet_scatter_energy_file_size_raw} and ~\ref{fig:alexNet_MobileNet_scatter_energy_image_size_raw}, the energy consumption refers solely to the original (raw) images from the Imagenet dataset. Figures ~\ref{fig:alexNet_mobileNet_scatter_energy_file_size_extended} and ~\ref{fig:alexNet_mobileNet_scatter_energy_image_size_extended} present the energy consumption results for the resized images. Finally, Figures ~\ref{fig:alexNet_mobileNet_scatter_energy_file_size_resized} and~ \ref{fig:alexNet_mobileNet_scatter_energy_image_size_resized} showcase the total energy consumption considering both the original and resized (extended) images.

The architectural differences between AlexNet and MobileNet play a significant role in their energy consumption profiles. MobileNet's lightweight design, which employs depthwise separable convolutions, reduces the number of operations required per layer, resulting in lower energy usage. In contrast, AlexNet's traditional convolutional layers demand more computations, leading to higher energy consumption. This comparison underscores the importance of optimizing network architectures for energy efficiency, particularly in resource-constrained environments.

Based on these results, it can be quickly concluded that MobileNet demonstrates superior performance to AlexNet regarding energy consumption, regardless of the type of image used (raw, resized, or extended).

\subsubsection{Correlation Analysis: Energy vs Image Size}

Table~\ref{tab:corr-alexnet-mobilenet} presents a high correlation between image size (height x width), file size, and the calculated energy for the classification process. As there was a high correlation between the two parameters, we chose to focus on image size, as it generally performed better as a way of simplifying the presentation of the results. 

The analysis of the warm-up stage (first 50 images) versus the steady-state stage (next 50 images) reveals distinct energy consumption profiles. During the warm-up, the system stabilizes its internal parameters, resulting in a slightly higher energy consumption compared to the steady-state stage. This stabilization period is crucial for ensuring fair comparisons across networks, as it eliminates transient effects that could distort results. By understanding this behavior, we can better interpret the energy efficiency of each neural network under realistic operating conditions.

\begin{table}[htbp]
\caption{\texttt{AlexNet} (A) and \texttt{MobileNet} (M) Energy Correlations. The column \texttt{raw} refers to the original image from the \texttt{ImageNet} dataset, and the column \texttt{extended} refers to the union between the original and \texttt{resized} image.}
\centering
\begin{tabular}{ |c|c|c|c|c| } 
\hline
& raw (A) & ext. (A) & raw (M) & extd (M)\\ \hline
file size  & 0.780582  & 0.776900 & 0.837781 &  0.836945\\ \hline
image size & 0.975476  & 0.974643 & 0.861603 &  0.828577 \\ \hline
\end{tabular}
\label{tab:corr-alexnet-mobilenet}
\end{table}

\subsubsection{Scatter AlexNet Results}
\label{sec:alexnet-scatter-results}
Figure~\ref{fig:alexNet-scatter-linear-model-raw} and Figure~\ref{fig:alexNet-scatter-linear-model-extended} refer to the energy median of heated stage scatter per image size. Figure~\ref{fig:alexNet-scatter-linear-model-raw} presents data from the dataset's $22.948$ original images called \texttt{raw}. Figure~\ref{fig:alexNet-scatter-linear-model-extended} gives the $2* 22.948 = 45.896$, which refers to the union between the original data image and resized images, called \texttt{extended}.

Both data distributions approximate a line. This aspect makes linear methods possible since the relationship between the image size feature and the energy target is expressed through an equation obtained from a linear regression.

\subsubsection{AlexNet Linear Model}
\begin{figure}[htbp]
\centering
\caption{Energy consumption distribution for AlexNet with raw images, highlighting the fitted linear regression model.}
\includegraphics[width=\columnwidth]{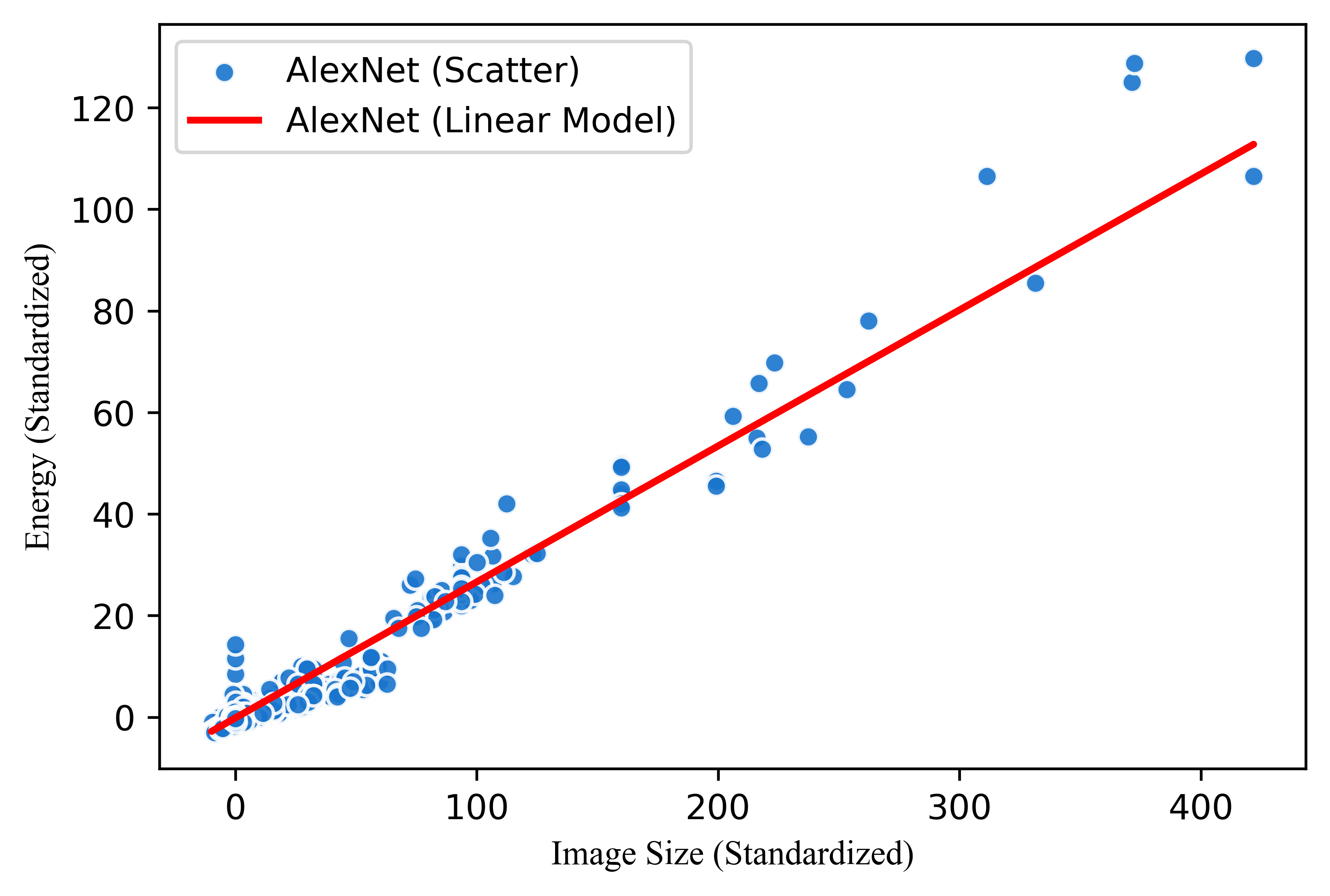}
\label{fig:alexNet-scatter-linear-model-raw}
\end{figure}

\begin{figure}[htbp]
\centering
\caption{Energy consumption distribution for AlexNet with the extended dataset, modeled using linear regression.}
\includegraphics[width=\columnwidth]{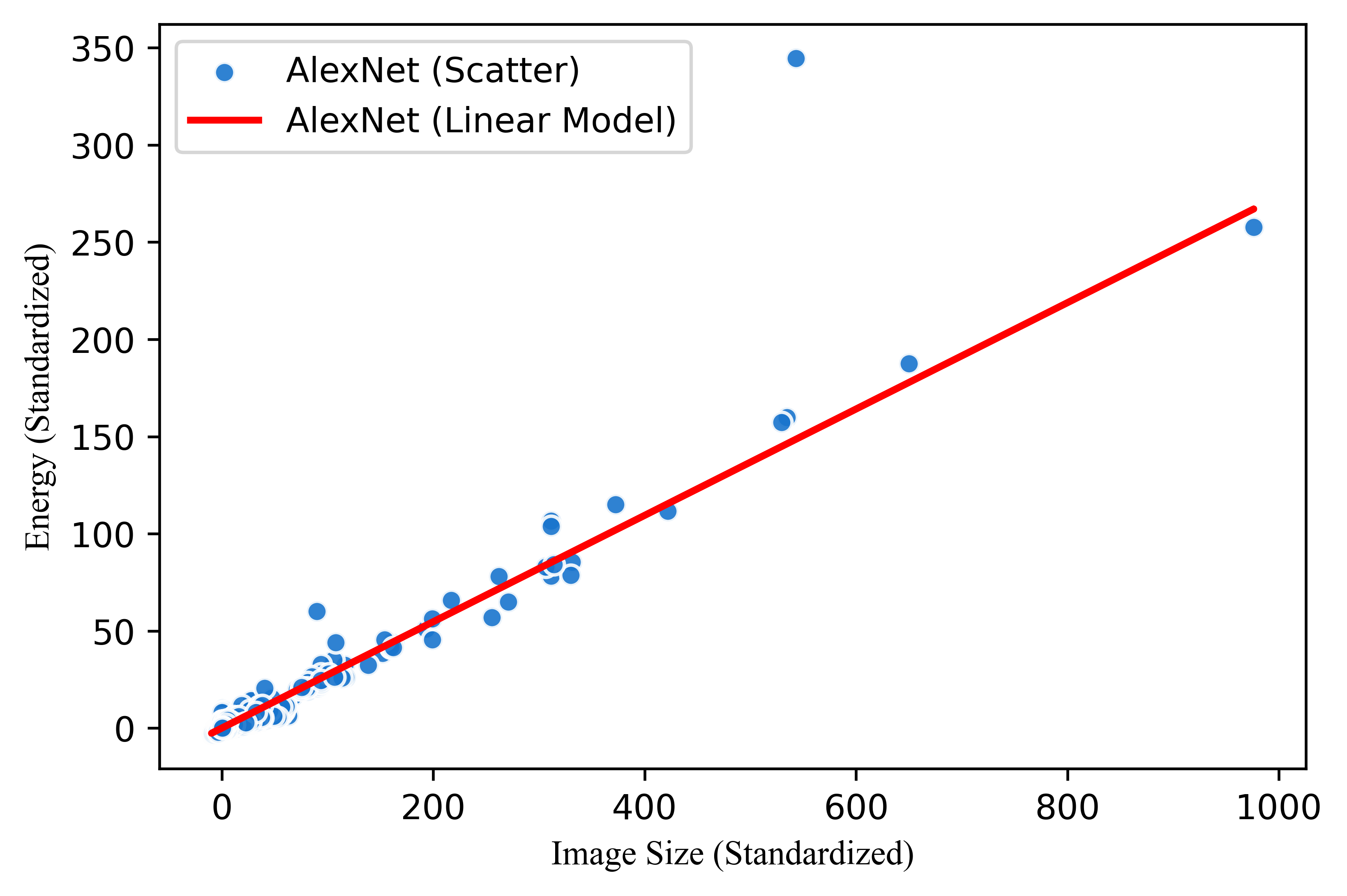}
\label{fig:alexNet-scatter-linear-model-extended}
\end{figure}

Figures~\ref{fig:alexNet-scatter-linear-model-raw} and~\ref{fig:alexNet-scatter-linear-model-extended} present the energy distributions by image size and the respective lines described by the linear models obtained on these. 
These linear models were obtained by the \texttt{LinearRegression} class from the \texttt{Python} \texttt{sklearn} linear model implementation. The methodology for obtaining it is based on three steps:

\begin{enumerate}
    \item Split the dataset into two sets, one for training and one for testing. A common practice is to set aside 30\% for testing and the rest for training;
    \item The model is trained with the training data obtained in the previous step. The training takes place through the ordinary least squares linear regression to minimize the residual sum of squares between the observed targets in the dataset and the targets predicted by the linear approximation. At the end of the process, two values result in the independent and dependent terms;
    \item The model is tested with the test set and obtained from which the mean square error can be obtained, which is widely used to assess the quality of the model obtained.

\end{enumerate}

The line of Figure~\ref{fig:alexNet-scatter-linear-model-raw} is represented by equation~\ref{eq:alexnet-raw-line-equation}, whose mean square error is $1.994687e-03$. On the other hand, the line of Figure~\ref{fig:alexNet-scatter-linear-model-extended} is represented by equation~\ref{eq:alexnet-extended-line-equation}, whose mean square error is $1.745484e-03$.

\begin{equation}
    \label{eq:alexnet-raw-line-equation}
    y= 0.36372127750480765 + 5.8291836e-07*x
\end{equation}

\begin{equation}
    \label{eq:alexnet-extended-line-equation}
    y= 0.3582136095619409 + 5.94579211e-07*x
\end{equation}

\begin{figure}[htbp]
\centering
\caption{Comparison of linear models for raw and extended datasets in AlexNet, illustrating close alignment.}
\includegraphics[width=\columnwidth]{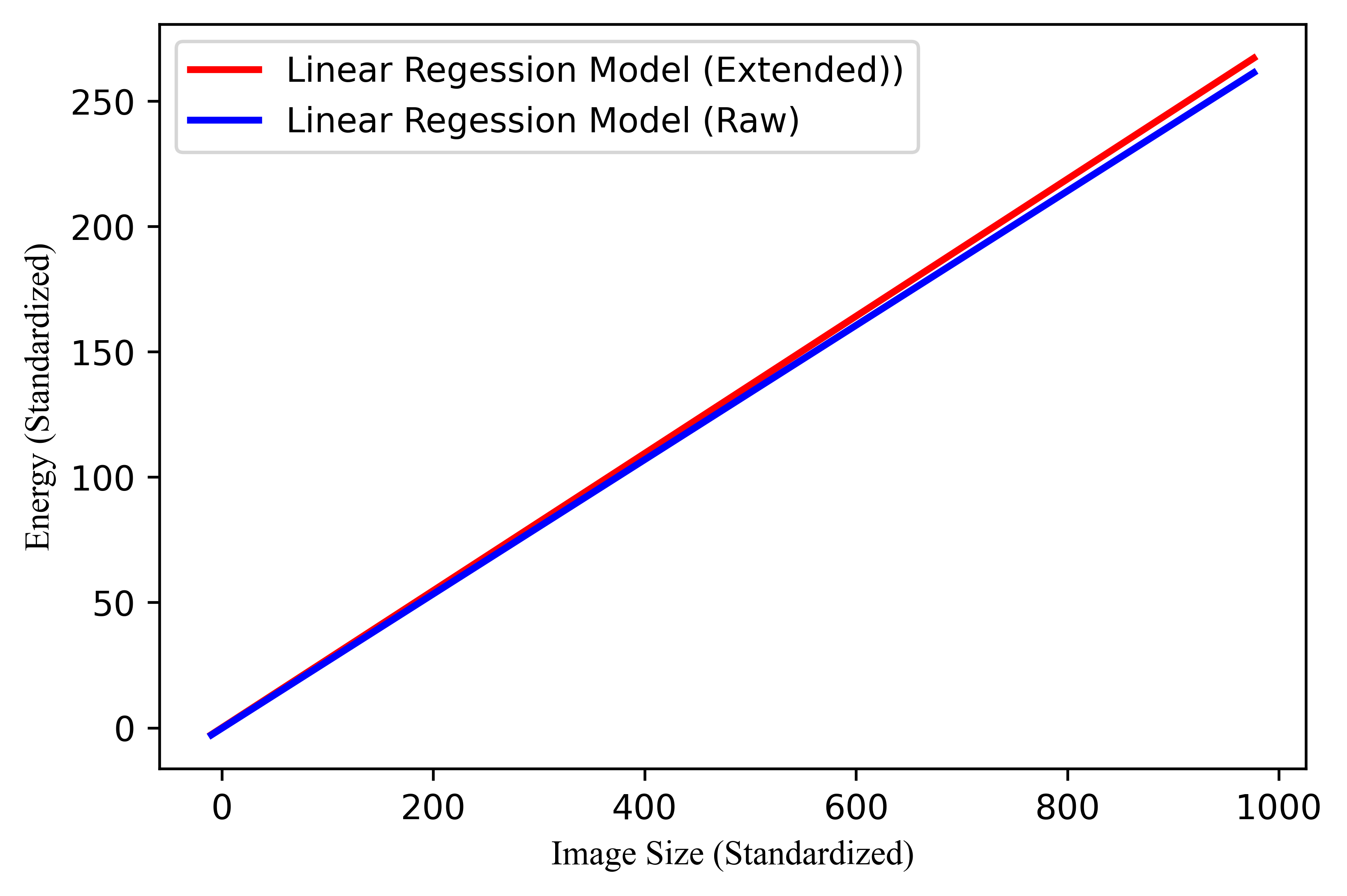}
\label{fig:alexNet-linear-model-raw-extended}
\end{figure}

Figure~\ref{fig:alexNet-linear-model-raw-extended} presents the two linear models, the raw and extended of the \texttt{AlexNet}. The lines are very close, almost equal for both dependent terms equations.

\subsubsection{Scatter \texttt{MobileNet} Results}
Figures~\ref{fig:mobileNet-scatter-linear-model-raw} and~\ref{fig:mobileNet-scatter-linear-model-extended} refer to data obtained from \texttt{MobileNet} network executions. As in Section~\ref{sec:alexnet-scatter-results}, the same original images were used for the results. However, the reduced images may differ since the most correctly classified images may differ between networks. Both data distributions also approach a line, as in the distributions presented in Section 3. The next Section follows the same path as in Section~\ref{sec:alexnet-scatter-results}.

\subsubsection{MobileNet Linear Model}
%

\begin{figure}[htbp]
\centering
\caption{Energy consumption patterns for MobileNet with raw images, analyzed using regression models.}
\includegraphics[width=\columnwidth]{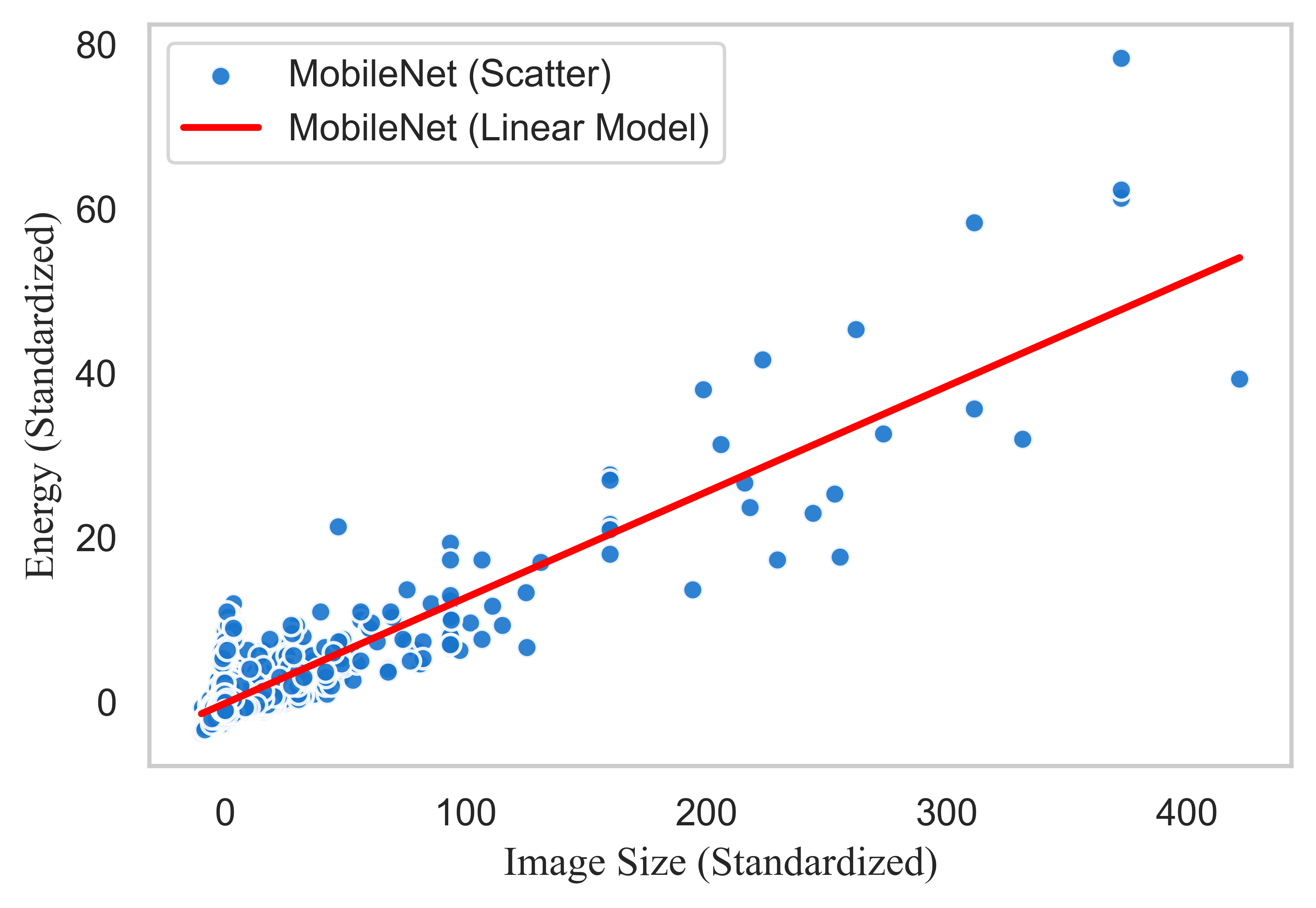}
\label{fig:mobileNet-scatter-linear-model-raw}
\end{figure}

\begin{figure}[htbp]
\centering
\caption{Energy consumption analysis for MobileNet on extended datasets, with linear regression modeling.}
\includegraphics[width=\columnwidth]{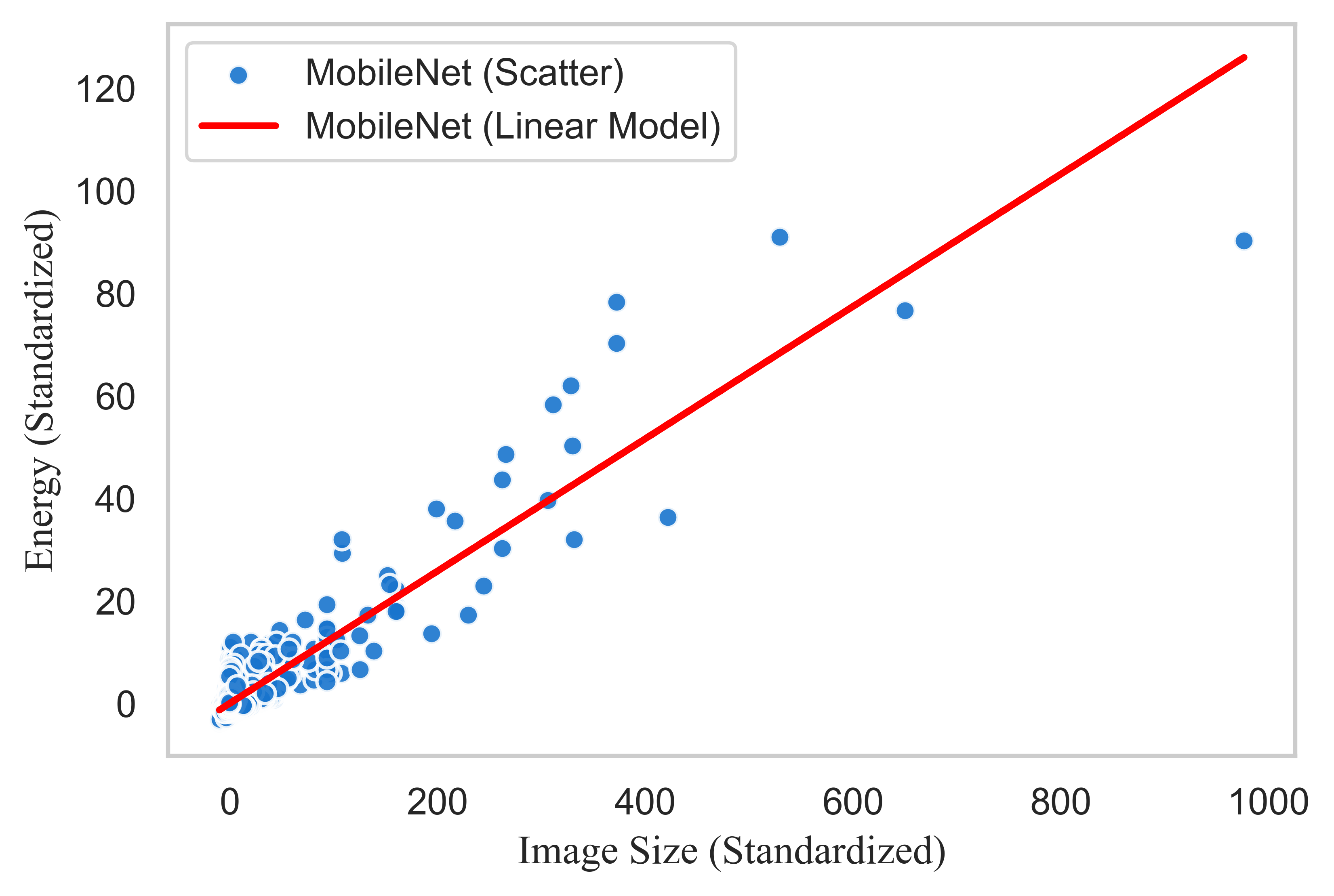}
\label{fig:mobileNet-scatter-linear-model-extended}
\end{figure}

The line of Figure~\ref{fig:mobileNet-scatter-linear-model-raw} is represented by equation~\ref{eq:mobilenet-raw-line-equation}, whose mean square error is $1.652666e-03$. On the other hand, the line of Figure~\ref{fig:mobileNet-scatter-linear-model-extended} is represented by equation~\ref{eq:mobilenet-extended-line-equation}, whose mean square error is $2.159426e-03$.

\begin{equation}
    \label{eq:mobilenet-raw-line-equation}
    y= 0.4066094107859453 + 2.07404002e-07*x
\end{equation}

\begin{equation}
    \label{eq:mobilenet-extended-line-equation}
    y= 0.38294462392290557 + 2.44975382e-07*x
\end{equation}

\begin{figure}[htbp]
\centering
\caption{Comparison of energy usage trends for raw and extended datasets in MobileNet.}
\includegraphics[width=\columnwidth]{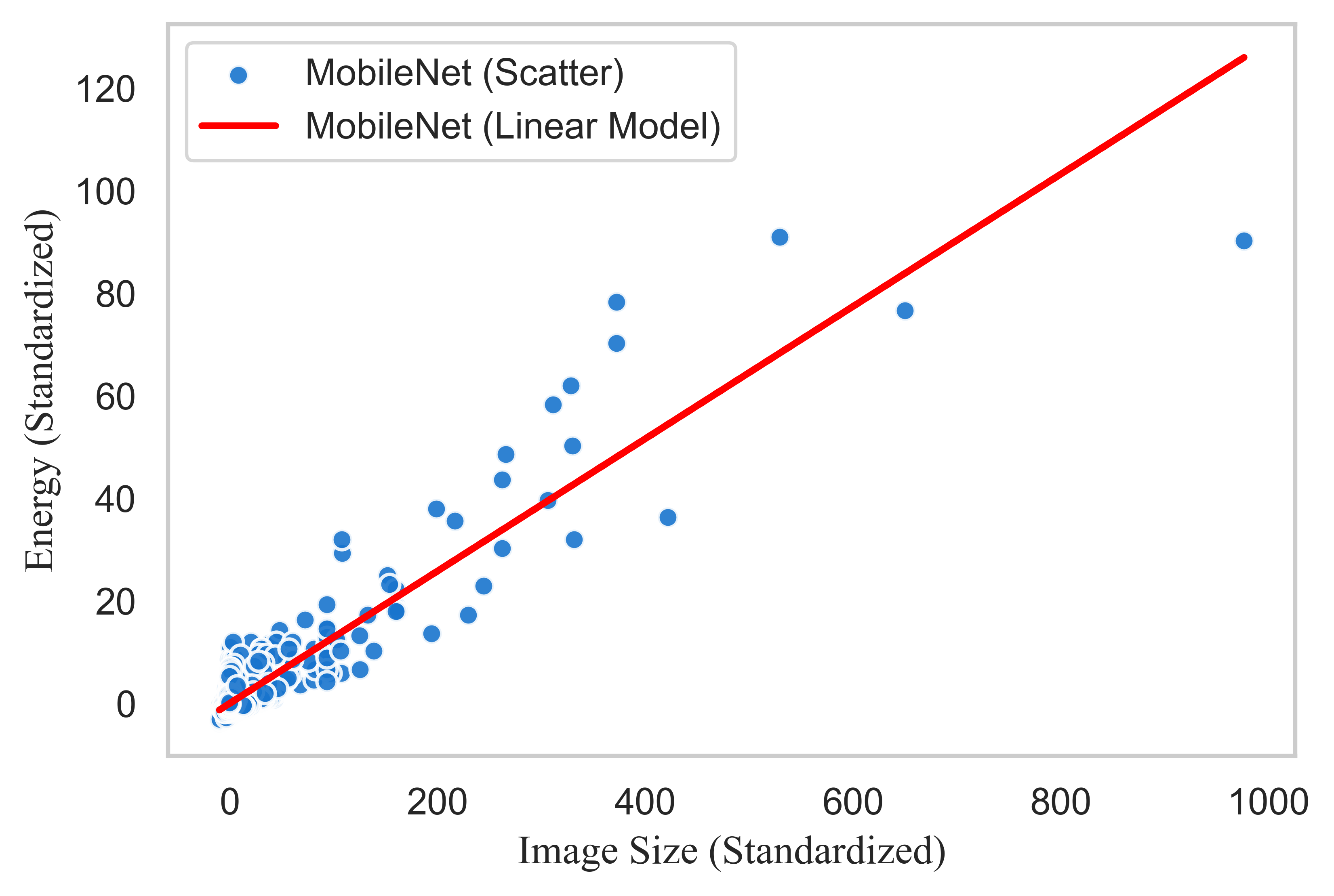}
\label{fig:mobileNet-linear-model-raw-extended}
\end{figure}

Figure~\ref{fig:mobileNet-linear-model-raw-extended} presents the two linear models, the raw and extended of the \texttt{MobileNet}. The lines are close, less than \texttt{AlexNet} models.

\begin{figure}[htbp]
\centering
\caption{Energy efficiency comparison between AlexNet and MobileNet using extended datasets.}
\includegraphics[width=\columnwidth]{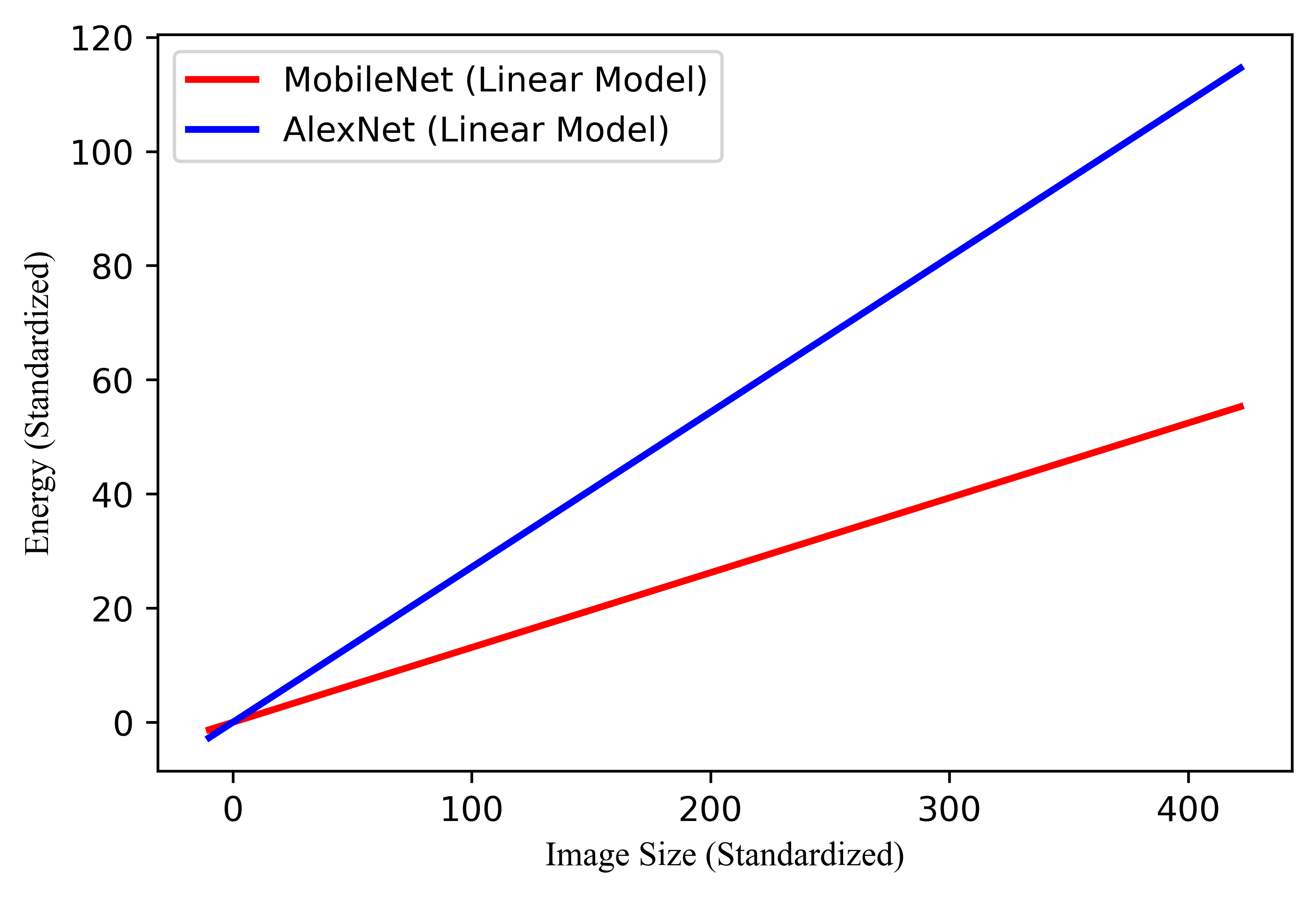}
\label{fig:alexnet-mobileNet-linear-model-raw-extended}
\end{figure}

Finally, Table~\ref{tab:alexnet-mobilenet-energy-median-std} shows the median energy consumption of \texttt{AlexNet} and \texttt{MobileNet} for the set of original and extended images. \texttt{MobileNet} proves to be more energy efficient than \texttt{AlexNet} in both sets. 2.32\% in extended and 6.25\%.
Figure~\ref{fig:alexnet-mobileNet-linear-model-raw-extended} shows the comparison between the \texttt{AlexNet} and \texttt{MobileNet} models. This vision reinforces \texttt{MobileNet's} energy efficiency over \texttt{AlexNet}.
In the next Section, we will see that it is not enough to look at a single perspective to choose a network application.

\begin{table}[htbp]
\caption{Comparison of median energy and standard deviation for AlexNet and MobileNet under different datasets.}
\centering
\begin{tabular}{ |c|c|c| } 
\hline
& Median Energy & Std. Dev. Energy\\ \hline
AlexNet Raw & 0.48 & 0.179665  \\ \hline
MobileNet Raw & 0.45 & 0.089804 \\ \hline
AlexNet Extended & 0.43  & 0.221197  \\ \hline
MobileNet Extended  & 0.42  &  0.080694 \\ \hline
\end{tabular}
\label{tab:alexnet-mobilenet-energy-median-std}
\end{table}

\subsubsection{Trade-off Accuracy}

\begin{figure*}
\centering
\caption{Trade-off analysis for AlexNet, showing the relationship between energy savings, accuracy, image size, and file size.}
\includegraphics[width=\textwidth]{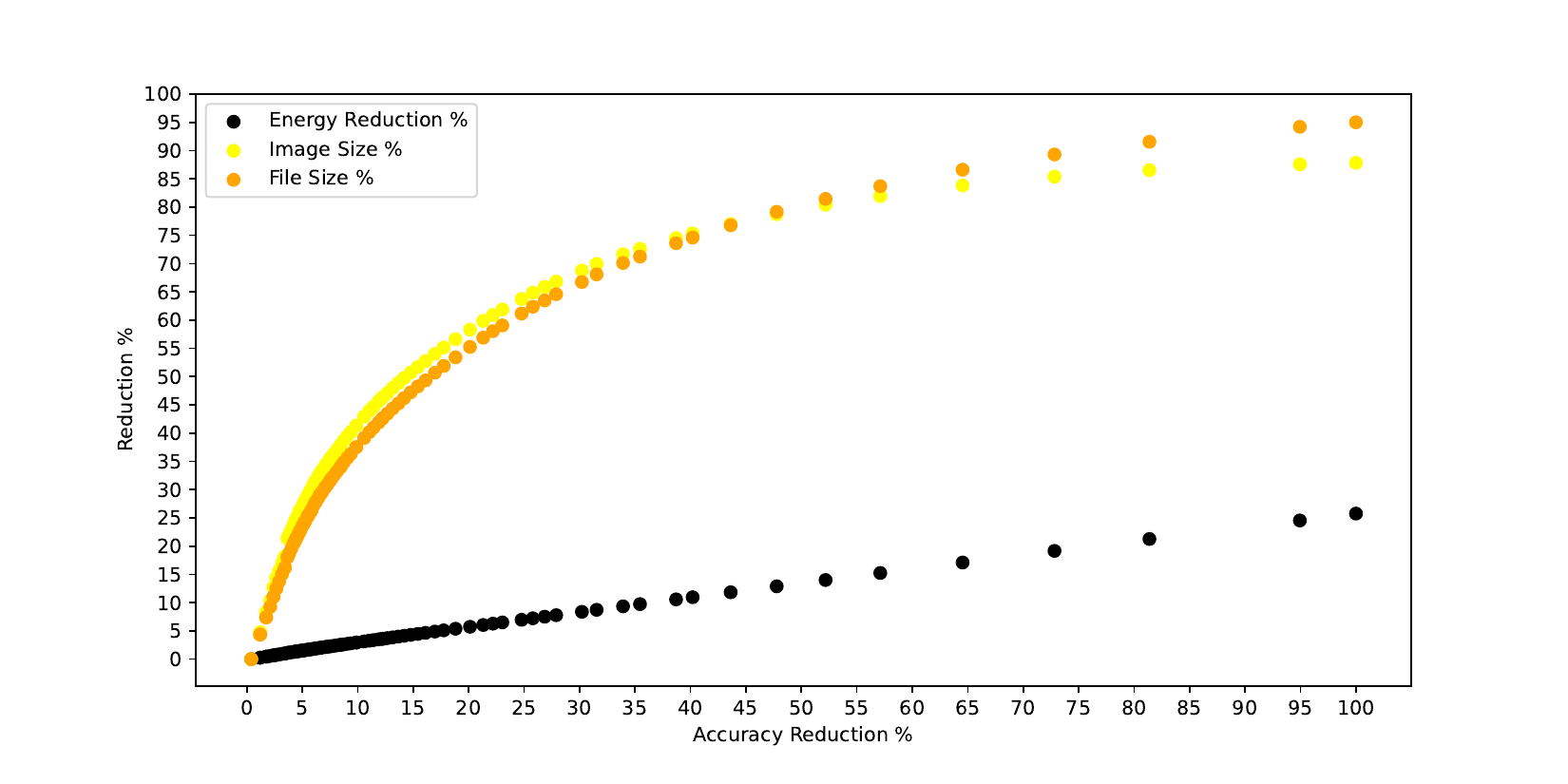}
\label{fig:alexNet-tradeoff}
\end{figure*}

\begin{figure*}
\centering
\caption{Trade-off analysis for MobileNet, emphasizing energy savings versus accuracy and data dimensions.}
\includegraphics[width=\textwidth]{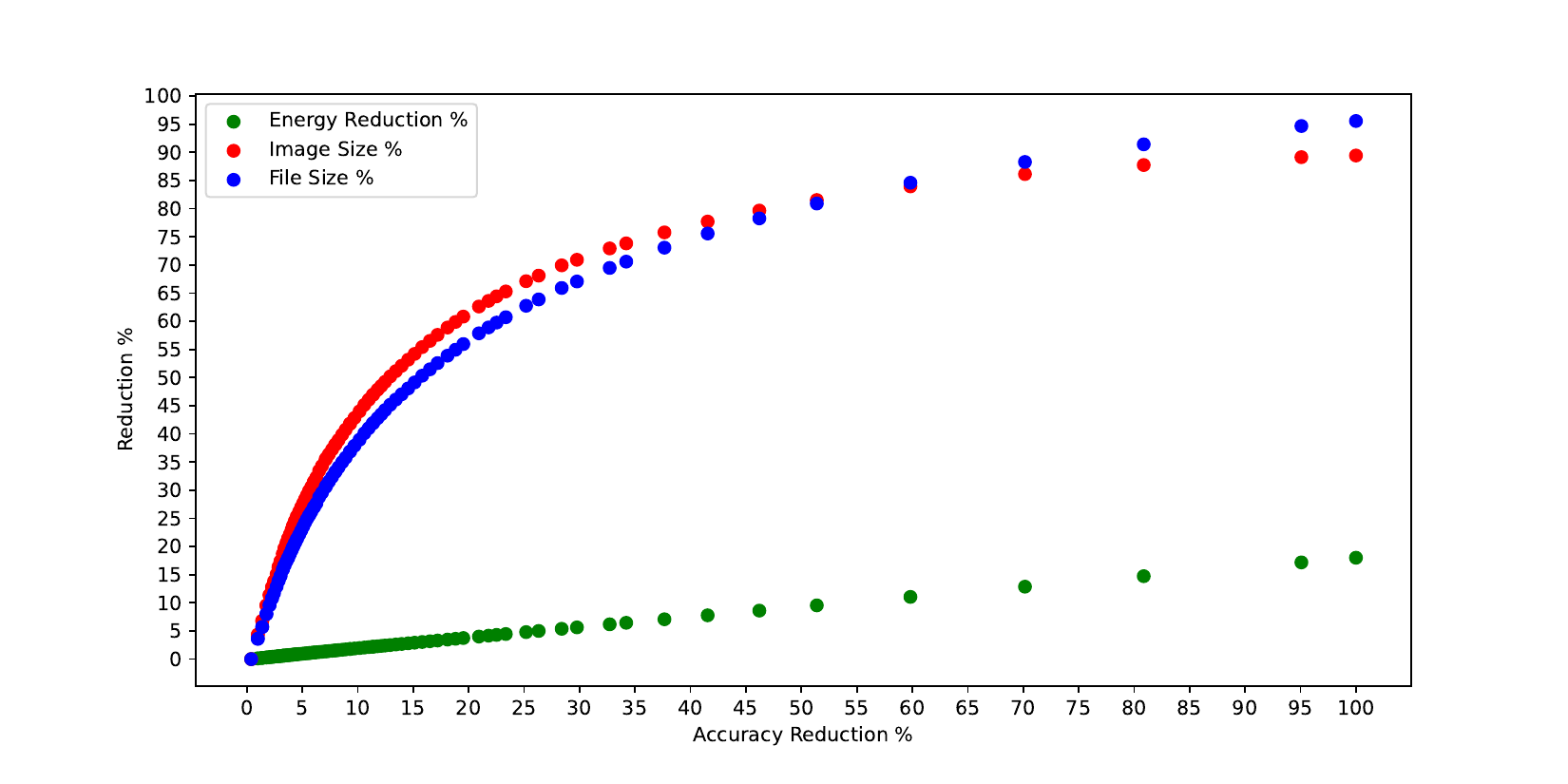}
\label{fig:mobileNet-tradeoff}
\end{figure*}

Figures~\ref{fig:alexNet-tradeoff} and~\ref{fig:mobileNet-tradeoff} address the accuracy trade-off~\cite{trade-off-accuracy-energy2021} by relating the percentage loss of accuracy as a function of the percentage loss of energy, image, and file size. Table~\ref{tab:percentage-energy-reduction-tradeoff-accuracy} was created with some data in these Figures to present them better. It can be seen that both networks lose some accuracy after the first reduction. Obtain approximately 1\% decrease in energy consumption, roughly $3\%$ loss in \texttt{AlexNet}, and $5\%$ in \texttt{MobileNet} accuracy. The other reductions show that \texttt{AlexNet} performs better than \texttt{MobileNet} since there is a substantial loss of accuracy for the same approximate decrease in energy consumption. Therefore, to achieve better energy-saving results, \texttt{MobileNet} must reduce images significantly, compromising the network’s accuracy.

\begin{table}[htbp]
\caption{Energy reduction and accuracy trade-off percentages for AlexNet and MobileNet under different configurations.}
\centering
\begin{tabular}{|c|l|rl|r|r|}
\hline
\multicolumn{1}{|l|}{energy} & DNN       & \multicolumn{2}{l|}{accuracy}   & \multicolumn{1}{l|}{file size} & \multicolumn{1}{l|}{image size} \\ \hline
\multirow{2}{*}{0\%}         & \texttt{AlexNet}   & \multicolumn{2}{r|}{0.40\%}     & 0.0\%                          & 0.0\%                           \\ 
                             & \texttt{MobileNet} & \multicolumn{2}{r|}{0.39\%}     & 0.0\%                          & 0.0\%                           \\ \hline
\multirow{2}{*}{$\sim$1\%}   & \texttt{AlexNet}   & \multicolumn{2}{r|}{$\sim$3\%}  & $\sim$16\%                     & $\sim$18\%                      \\  
                             & \texttt{MobileNet} & \multicolumn{2}{r|}{$\sim$5\%}  & $\sim$24\%                     & $\sim$28\%                      \\ \hline
\multirow{2}{*}{$\sim$2\%}   & \texttt{AlexNet}   & \multicolumn{2}{r|}{$\sim$6\%}  & $\sim$29\%                     & $\sim$32\%                      \\  
                             & \texttt{MobileNet} & \multicolumn{2}{r|}{$\sim$10\%} & $\sim$40\%                     & $\sim$45\%                      \\ \hline
\multirow{2}{*}{$\sim$3\%}   & \texttt{AlexNet}   & \multicolumn{2}{r|}{$\sim$10\%} & $\sim$39\%                     & $\sim$42\%                      \\  
                             & \texttt{MobileNet} & \multicolumn{2}{r|}{$\sim$16\%} & $\sim$55\%                     & $\sim$50\%                      \\ \hline
\multirow{2}{*}{$\sim$4\%}   & \texttt{AlexNet}   & \multicolumn{2}{r|}{$\sim$14\%} & $\sim$46\%                     & $\sim$49\%                      \\  
                             & \texttt{MobileNet} & \multicolumn{2}{r|}{$\sim$20\%} & $\sim$62\%                     & $\sim$58\%                      \\ \hline
\multirow{2}{*}{$\sim$5\%}   & \texttt{AlexNet}   & \multicolumn{2}{r|}{$\sim$18\%} & $\sim$52\%                     & $\sim$55\%                      \\  
                             & \texttt{MobileNet} & \multicolumn{2}{r|}{$\sim$28\%} & $\sim$66\%                     & $\sim$70\%                      \\ \hline
\multirow{2}{*}{$\sim$10\%}  & \texttt{AlexNet}   & \multicolumn{2}{r|}{$\sim$39\%} & $\sim$74\%                     & $\sim$74\%                      \\  
                             & \texttt{MobileNet} & \multicolumn{2}{r|}{$\sim$51\%} & $\sim$81\%                     & $\sim$82\%     \\ \hline                 
\end{tabular}
\label{tab:percentage-energy-reduction-tradeoff-accuracy}
\end{table}

\section{Case Study 2: Impact of Image File Formats on Energy Consumption}
\label{sec:case-study-2}

\begin{figure*}[h!]
\centering
\caption{Illustrates the automated workflow implemented by Phoeni6 to conduct energy evaluations for diverse neural network configurations and datasets. The diagram emphasizes the modularity of the framework, including dynamic dataset generation, conversion processes, and energy monitoring steps. This workflow ensures reproducibility and adaptability for varying experimental needs.}
\includegraphics[width=\textwidth]{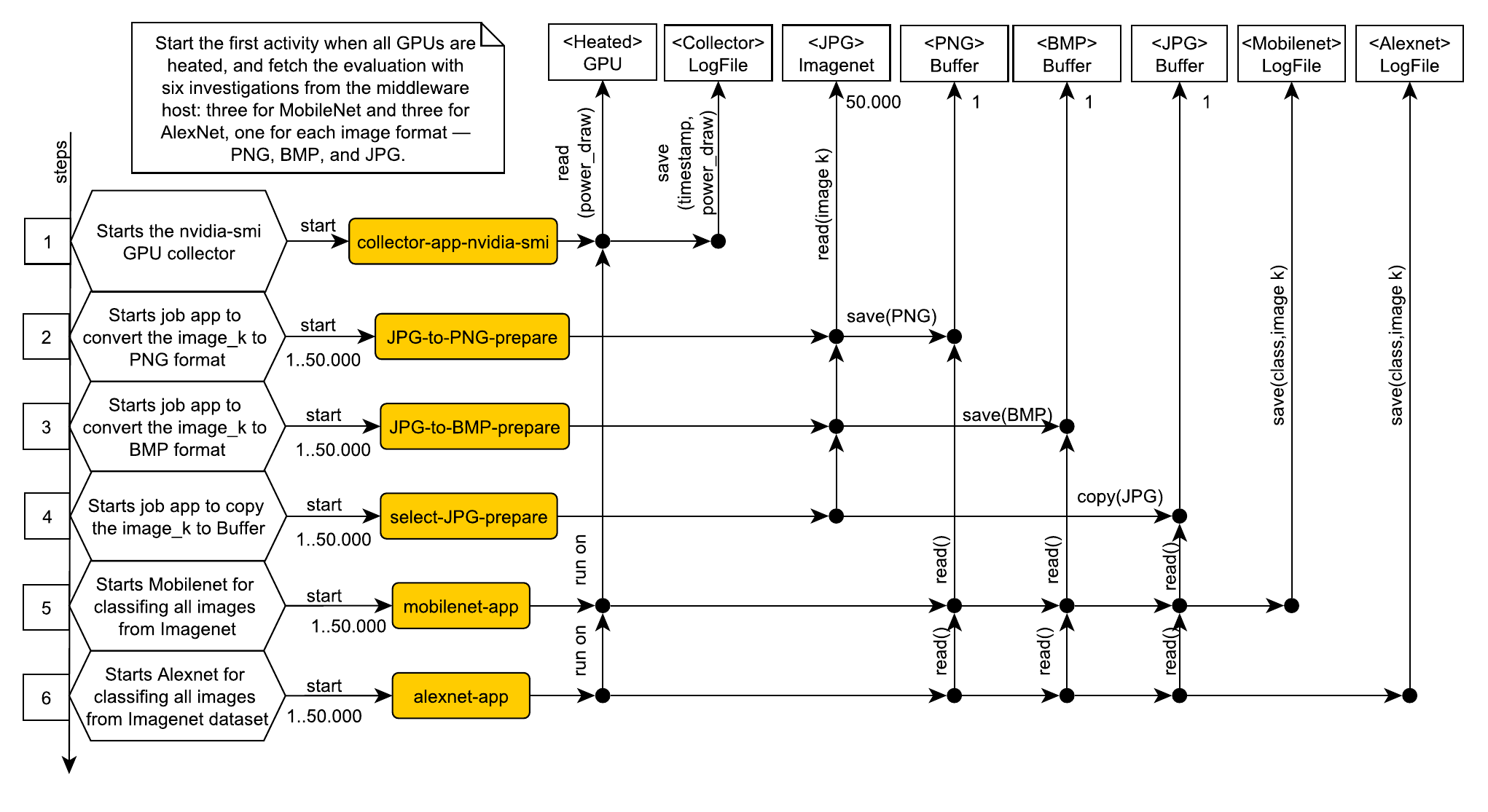}
\label{fig:case_study_2}
\end{figure*}

This section presents a second study that evaluates the energy consumption of the AlexNet and MobileNet networks using Phoeni6, focusing on the impact of image file formats on energy usage. The networks are tested while classifying 50,000 images from the ImageNet dataset, originally in JPG format, which are then converted to PNG and BMP for comparison. Figure~\ref{fig:case_study_2} provides an activity diagram illustrating how Phoeni6 addresses this issue, including the management of image copies after conversion. Similar to the first study presented in Section~\ref{sec:case-study}, this study also considers disk space usage due to the large number of image copies. The job-apps \texttt{JPG-to-PNG-prepare} and \texttt{JPG-to-BMP-prepare} alternate between image conversion and network execution to ensure flexibility.

This study further demonstrates the flexible and adaptive nature of Phoeni6 in tackling various aspects of the same problem, particularly the relationship between image formats and energy consumption. In addition to energy analysis, it offers a comparison between AlexNet and MobileNet, as introduced in the first study, by implementing a set of four job-app applications.


\begin{itemize}

    \item \textbf{select-JPG-prepare}: this application selects each original dataset image for investigation.

    \item \textbf{JPG-to-BMP-prepare}: this application converts each original JPG file from the ImageNet dataset to BMP format.

    \item \textbf{JPG-to-PNG-prepare}: this application converts each original JPG file format to PNG format for every dataset image.    


\end{itemize}

As in the first study, these job-apps were developed to manage the disk space used by all copies of files converted to BMP and PNG formats. At the end of each investigation, each newly generated file is deleted to free up space.

\begin{figure}[h!]
\centering
\caption{Energy consumption contributions for AlexNet and MobileNet across different image formats (PNG, JPG, BMP).}
\includegraphics[width=\columnwidth]{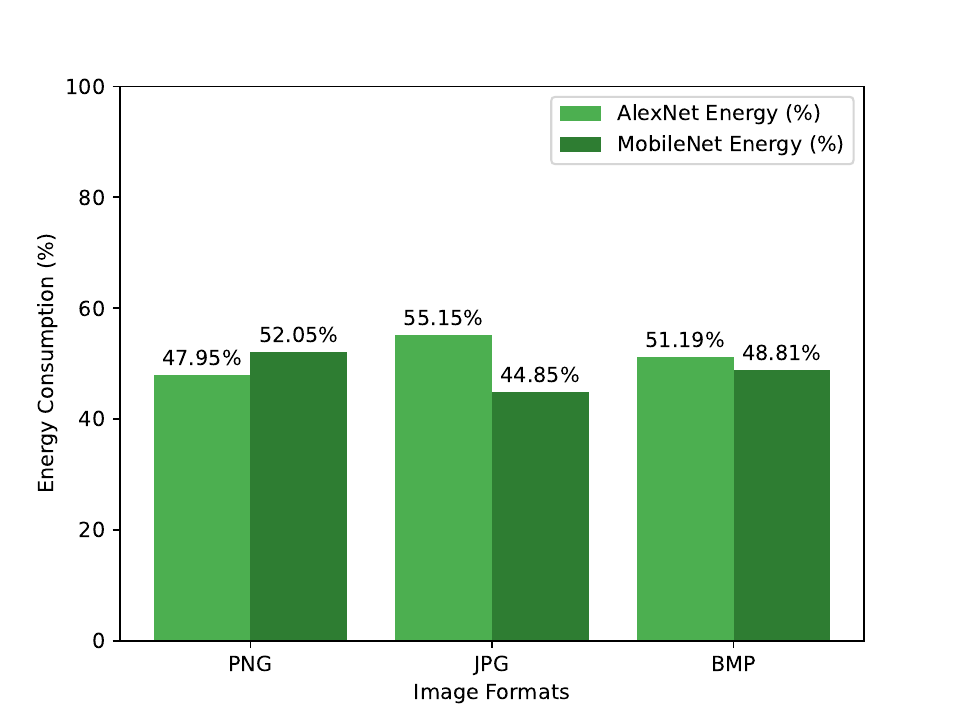}
\label{fig:alexnet-mobilenet-av-energy-gpu-0-formats}
\end{figure}

\begin{figure}[h!]
\centering
\caption{Energy consumption breakdown by image format, emphasizing the efficiency of BMP over PNG and JPG.}
\includegraphics[width=\columnwidth]{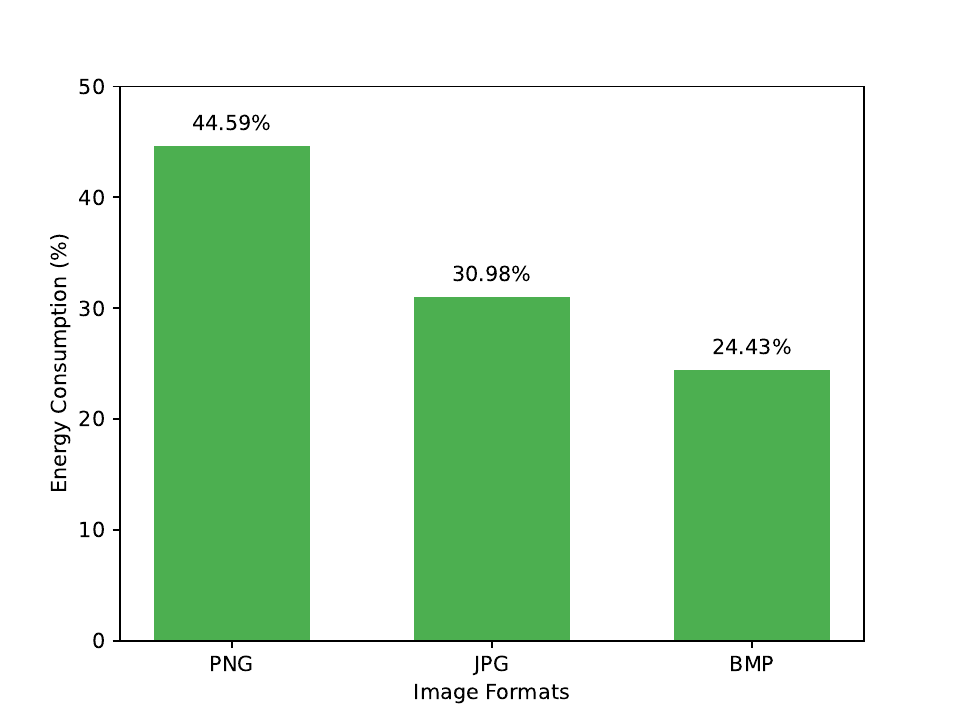}
\label{fig:total_energy_percentage-formats}
\end{figure}

\subsection{Evaluation Setup}
\label{sec:evaluation-setup}

To ensure a clear and structured understanding of the evaluation setup, the key components and configurations are summarized in tables. Table \ref{tab:evaluation-objectives} outlines the main objectives and guiding questions of the evaluation. Table \ref{tab:hardware-configuration} details the hardware specifications used for the experiments. Table \ref{tab:dataset-configuration} describes the dataset properties and parameters evaluated. Finally, Table \ref{tab:application-configuration} highlights the applications employed to manage, collect, and analyze energy consumption data. This tabular representation provides a concise yet comprehensive view of the evaluation setup.

\begin{table*}[h!]
\centering
\caption{Evaluation Objectives and Key Questions}
\label{tab:evaluation-objectives}
\begin{tabular}{|p{4cm}|p{8cm}|}
\hline
\textbf{Objective}          & \textbf{Key Questions}                                                             \\ \hline
Energy Efficiency Analysis  & How does the energy consumption vary across different image formats and models?    \\ \hline
Model Performance Validation & Can the framework handle diverse datasets and ensure reproducibility of results?  \\ \hline
Hardware Scalability         & How does the setup adapt to diverse hardware configurations?                      \\ \hline
\end{tabular}
\end{table*}

\begin{table*}[h!]
\centering
\caption{Hardware Configuration for Evaluation}
\label{tab:hardware-configuration}
\begin{tabular}{|p{6cm}|p{6cm}|}
\hline
\textbf{Component}        & \textbf{Specification}                                   \\ \hline
CPU                      & Intel Core i7-9700 @ 3.00GHz                             \\ \hline
GPU                      & NVIDIA GTX 1080 Ti with 11GB GDDR5X                     \\ \hline
RAM                      & 32GB DDR4                                               \\ \hline
Storage                  & 1TB SSD                                                 \\ \hline
Operating System         & Ubuntu 20.04 LTS                                        \\ \hline
\end{tabular}
\end{table*}

\begin{table*}[h!]
\centering
\caption{Dataset Configuration}
\label{tab:dataset-configuration}
\begin{tabular}{|p{6cm}|p{6cm}|}
\hline
\textbf{Parameter}        & \textbf{Details}                                        \\ \hline
Dataset Name             & ImageNet                                                \\ \hline
Formats Evaluated        & JPG, BMP, PNG                                           \\ \hline
Number of Images         & 150,000 (50,000 per format)                             \\ \hline
Image Resolution         & 224x224 pixels                                          \\ \hline
\end{tabular}
\end{table*}

\begin{table*}[h!]
\centering
\caption{Applications Used in Evaluation}
\label{tab:application-configuration}
\begin{tabular}{|p{4cm}|p{8cm}|}
\hline
\textbf{Application}      & \textbf{Purpose}                                        \\ \hline
Manager-app              & Orchestrates containers and manages evaluation workflow \\ \hline
Collector-app            & Collects power consumption data from hardware          \\ \hline
Analyzer                 & Processes logs and generates energy efficiency metrics \\ \hline
\end{tabular}
\end{table*}

\subsection{Results}

The following subsections present a detailed analysis of the results, highlighting the energy consumption across different image formats, specific observations about the models and formats, and a summary of key findings. This structured approach aims to provide deeper insights into how energy efficiency is influenced by neural network architectures and data formats.

\subsubsection{Consolidated Results}

~\ref{fig:alexnet-mobilenet-av-energy-gpu-0-formats} and~\ref{fig:total_energy_percentage-formats} show the consolidated results after 150,000 investigations, comprising 50,000 images in the original JPG format, 50,000 in the JPG-to-BMP converted format, and another 50,000 in the JPG-to-PNG converted format. The energy consumption was calculated using the average power measured at regular intervals during the inference process. These metrics provide a quantitative basis for comparing the computational cost of different file formats and neural network architectures, ensuring a reproducible and fair comparison.

\subsubsection{Energy Consumption Across Formats}

The results indicate that both models consume significantly more energy when processing images in PNG format, with AlexNet being slightly more energy-intensive than MobileNet. For images in JPG and BMP formats, energy consumption is considerably lower, particularly for BMP, which exhibits the lowest energy demand for both networks. Specifically, for PNG images, the average energy consumption per classification was 15\% higher for AlexNet and 10\% higher for MobileNet compared to JPG images. This increase can be directly attributed to the overhead introduced by decoding the compressed PNG format. In contrast, BMP images required 20\% less energy than JPG on average, showcasing the impact of processing uncompressed formats. These data suggest that the image format directly influences the energy efficiency of classification models.

\subsubsection{Specific Observations}

\begin{itemize}
    \item \textbf{PNG vs. JPG and BMP Formats}: both models (AlexNet and MobileNet) show significantly higher energy consumption when processing PNG images, which suggests that this format may involve more complex processing requirements, possibly due to its higher complexity and the need for decoding or compression/decompression compared to JPG and BMP. The higher complexity of PNG leads to an average energy increase of 15\%, highlighting the non-linear relationship between compression complexity and energy demand.
    \item \textbf{Difference Between AlexNet and MobileNet}: AlexNet’s higher energy intensity compared to MobileNet for PNG images could be attributed to its more complex architecture and greater number of parameters, leading to a higher computational load. MobileNet consistently demonstrated lower energy consumption due to its lightweight architecture, achieving up to 6\% higher energy efficiency compared to AlexNet when processing PNG images. This suggests that architectural optimizations in MobileNet are effective even under higher computational loads.
    \item \textbf{BMP Format}: The BMP format, which shows lower energy demand, might be simpler to process due to its lack of compression and straightforward pixel data storage, resulting in reduced computational overhead for both models. The reduced computational overhead of BMP allows for a decrease of 20-30\% in energy consumption, presenting an opportunity for scenarios where storage constraints are not critical but energy efficiency is.
\end{itemize}

\subsubsection{Summary of Findings}

These findings highlight the necessity of scalable solutions, which will be explored further as outlined in Section~\ref{sec-future-work} Future Work, to address the challenges of large-scale neural networks and diverse hardware configurations.

These findings underscore the importance of considering image format when evaluating the energy efficiency of classification models. The findings reveal that energy consumption is strongly influenced by image formats, with BMP being the most energy-efficient and PNG the most demanding. If energy efficiency is a critical concern, optimizing image formats or adapting models to reduce energy consumption may be beneficial. Further analysis into per-layer energy consumption and model-specific optimization could provide deeper insights into these dynamics.

\section{Related works}
\label{sec:related-works}
In this section, a systematic review of papers and surveys was carried out to identify the main approaches related to the energy consumption of neural networks. The selection of articles was made through a search on Google Scholar using the related search strings below:

 \begin{itemize}
    \item ("Neural Networks" OR "Deep Learning") AND "Energy Consumption"
    \item ("Machine Learning" OR "Artificial Intelligence") AND ("Energy Consumption" OR "Power Consumption")
    \item ("GPU" OR "CPU") AND "Energy Efficiency"
    \item ("Neural Networks" OR "Deep Learning") AND ("Power Consumption" OR "Power Estimation")
    \item ("Artificial Intelligence" OR "Machine Learning") AND ("Energy Estimation" OR "Power Estimation")
\end{itemize}
The selection of the articles was made through the reading of the titles, summary, and, finally, the introduction. The selected papers were grouped into the following categories:
 \begin{itemize}
    \item Energy-measurement methods and tools
    \item Network compression approaches
    \item Training-parameter modification approaches
    \item Hardware-parameter modification methods
    \item Energy consumption studies without intervention
\end{itemize}

This Section places a greater emphasis on works falling under the category of Energy Measurement Methods and Tools, as their approach aligns more closely with Phoeni6. Five such works are subject to analysis and comparison with Phoeni6. While works in other categories maintain relevance due to their adaptability to the Phoeni6 model, only a single exemplary work from each category is spotlighted to showcase the potential of the Phoeni6 approach.

\subsection{Energy Measurement: Methods and Tools}

The works described in the following sub-subsections~\cite{liu2022energyprofiling,li2022neuralpower,chen2021powermeter,zhang2021energyefficient,wang2020systematic} address challenges similar to those addressed by Phoeni6, such as the need for reproducible and fair comparison of energy consumption results. They also provide different approaches to energy consumption evaluation, such as using hardware-aware power measurement techniques or developing frameworks for automated profiling. Table~\ref{tab:comparation-2} compares Phoeni6 and these works.

\begin{table*}[htbp]
\centering
\caption{Feature comparison between Phoeni6 and other frameworks, emphasizing advantages in portability and automation. EP = EnergyProfilin, NP = NeuralPower, PM = PowerMete.}
\begin{tabular}{|c|c|c|c|c|c|c|}
\hline
Feature & Phoeni6 & EP ~\cite{liu2022energyprofiling}& NP\cite{li2022neuralpower} & PM \cite{chen2021powermeter}& Energy-Efficient \cite{zhang2021energyefficient} & Systematic \cite{wang2020systematic}\\
\hline
Portability & Yes & Yes & Yes & Yes & No & No \\\hline
Automation & Yes & No & No & No & No & No \\\hline
Transparency & Yes & No & No & No & No & No \\\hline
Coordination & Yes & No & No & No & No & No \\\hline
Centralized DB & Yes & No & No & No & No & No \\\hline
Multiplatform & Yes & No & Yes & Yes & No & No \\\hline
Data granularity & Yes & No & Yes & Yes & No & No \\\hline
Complex analytical investigation & Yes & No & Yes & Yes & No & No \\\hline
General-purpose & Yes & No & Yes & Yes & Yes & Yes \\
\hline
\end{tabular}
\label{tab:comparation-2}
\end{table*}

\subsubsection{EnergyProfiling \cite{liu2022energyprofiling}} EnergyProfiling is a user-friendly framework for energy profiling of deep neural networks on mobile devices, offering tools to collect and analyze energy consumption data across various neural network models and hardware platforms. While it provides portability and ease of use, it lacks automated profiling, transparency, and coordination, requiring manual data collection and analysis, potentially leading to time-consuming and error-prone processes. Moreover, the absence of a centralized database for sharing energy consumption data hinders comparing findings across different studies.

\subsubsection{NeuralPower \cite{li2022neuralpower}} NeuralPower offers an efficient and accurate framework for automated power profiling of neural networks, employing hardware-aware power measurement techniques to gather data across various models and hardware platforms. Nevertheless, it lacks transparency and coordination, necessitating adaptation by researchers for specific use cases. Furthermore, the absence of a centralized database for sharing collected energy consumption data via NeuralPower poses challenges in comparing results across different studies.

\subsubsection{PowerMeter \cite{chen2021powermeter}} PowerMeter, a hardware-aware power measurement framework for deep neural networks, equips researchers with tools to collect and analyze power consumption data across a range of neural network models and hardware platforms. Its focus on accuracy and comprehensiveness renders it an ideal choice for in-depth power analysis. Nevertheless, PowerMeter lacks transparency and coordination, requiring adaptation for those aiming to assess energy consumption in mobile devices. Furthermore, the absence of a central database for sharing energy consumption data gathered through PowerMeter complicates the comparison of results among various research studies.

\subsubsection{Energy-Efficient Neural Network Design: A Survey \cite{zhang2021energyefficient}} The survey offers a comprehensive overview of energy-efficient neural network design techniques, encompassing model compression, pruning, and hardware acceleration, thus serving as a valuable resource for researchers and developers interested in designing energy-efficient neural networks. However, it falls short in providing tools or frameworks for directly assessing energy efficiency, placing the onus on researchers to develop their tools, a potentially time-consuming and challenging endeavor. Furthermore, the absence of a centralized database for sharing energy consumption data across various studies complicates the comparative analysis of results among different research groups.

\subsubsection{A Systematic Evaluation of Energy Efficiency of Neural Networks \cite{wang2020systematic}} The systematic evaluation assesses the energy efficiency of diverse neural network models and hardware platforms using various benchmarks and metrics, encompassing factors like inference time, throughput, and power consumption, yielding valuable insights into enhancing neural network energy efficiency. Nevertheless, it lacks a prescribed methodology for the systematic collection and analysis of energy consumption data, placing the onus on researchers to develop their own approaches, which can be laborious and prone to errors. Furthermore, the absence of a centralized database for sharing energy consumption data from the evaluation hampers comparing results across different studies.

\subsection{Additional Research Categories}
\label{subsec:other-categories}

Here, we present a representative example of one work from each category to illustrate how Phoeni6 can support a variety of approaches with relative ease.

\subsubsection{Network compression approaches} 
Gholam et al.~\cite{Survey-Quantization-2021} presents a comprehensive overview of quantization methods for efficient neural network inference, categorizing these methods into fixed-point, variable-point, and dynamic quantization, discussing the merits and drawbacks of each category and providing specific examples. Phoeni6 could enhance this work in several ways, such as collecting data on the energy consumption of different quantization methods, including fixed-point, variable-point, and dynamic quantization, across various neural networks and hardware platforms. This data would validate the claims in the article and identify the most energy-efficient quantization methods for diverse applications.
A few other works from our systematic survey fall in this category~\cite{zhou2016dorefa,gong2019differentiable,dotzel2023fliqs}.

\subsubsection{Training parameter modification approaches}
Brownlee et al.~\cite{trade-off-accuracy-energy2021} provide a comprehensive overview of the factors that affect the accuracy and energy consumption of machine learning models, focusing on neural networks. Phoeni6 supports this work through several approaches: it can collect data on the accuracy and energy consumption of different machine learning models on various hardware platforms, helping to validate claims and identify the most energy-efficient models for specific applications and platforms. 
In addition to the works mentioned above, our systematic survey identified the following works as relevant~\cite{patterson2021carbon,mehlin2023towards,metz2020tasks}.
    
\subsubsection{Hardware parameter modification methods}
Tang et al.~\cite{Impact-GPU-DVFS:2019} investigated the effects of GPU dynamic voltage and frequency scaling (DVFS) on the energy and performance of deep learning workloads.
Phoeni6 could establish a general framework for collecting, analyzing, and comparing DVFS impact data, promoting the development of new techniques for enhancing the energy and performance of deep learning workloads. Overall, Phoeni6 has the potential to be a valuable tool for supporting research related to DVFS and its impact on deep learning workloads by enabling data collection, tool development, and data sharing among researchers.
Our systematic survey also found the following works relevant~\cite{tang2019impact,mei2017survey,mei2013measurement}.

\subsubsection{Energy consumption studies without any intervention:} 
Castro et al.~\cite{Energy-based-tuning:2018} presented a novel framework for reducing the energy consumption of convolutional neural networks (CNNs) on multi-GPUs.
Phoeni6 can offer substantial support to this research. First, it can collect data on the energy consumption of different CNNs on various multi-GPU platforms, enabling validation of energy consumption models and identification of opportunities for energy reduction. Additionally, Phoeni6 can be used to develop tools that automate the energy-based tuning process for CNNs, considering layer-specific energy consumption, communication overhead, and accuracy thresholds, simplifying the utilization of the proposed framework for energy-efficient CNN tuning.
Other relevant works from our systematic survey include~\cite{yao2021evaluating,yao2021evaluating,sun2021evaluating,zhao2023sustainable}.

\subsubsection{Final Analysis:} 

Table~\ref{tab:comparation-2} highlights several key characteristics that underscore the relevance of Phoeni6 within its contextual framework. However, as discussed in Section~\ref{sec:methodology} and briefly examined through the case studies in Sections ~\ref{sec:case-study} and~\ref{sec:case-study-2}, initiating an evaluation requires prior configurations. These configurations demand that the researcher possess a thorough understanding of both the underlying concepts and the preparatory activities, such as:

\begin{itemize}
    \item Containerization of new networks to be used in the evaluations;
    \item Containerization of new collectors when they differ from those already containerized, as is the case with nvidia-smi;
    \item Proper identification of devices to be utilized, such as GPUs;
    \item A working knowledge of the database to facilitate the replication of evaluations.
\end{itemize}

\section{Conclusion}
\label{sec:conclusion}
This work introduces Phoeni6, a systematic approach aimed at evaluating the energy consumption of neural networks while adhering to the principles of fair comparison (FC) and result reproducibility (RR). As the importance of energy efficiency in neural networks becomes increasingly evident, Phoeni6 offers a robust solution for managing large volumes of energy-related data and configurations, ensuring portability, transparency, and coordination throughout the evaluation process.

Phoeni6 features a set of containerized tools for operating system portability, a database management system (DBMS) for data persistence, and a comprehensive data model for manipulating and storing information. These components streamline the configuration, monitoring, and management of energy consumption evaluations. The methodology employed by Phoeni6 automates the evaluation process and addresses specific questions related to monitoring tools for various devices, data formats, and data persistence.

A comparative case study involving two different neural networks validates the effectiveness of Phoeni6, demonstrating its value in guiding energy consumption evaluations. This research contributes to the global emphasis on energy-efficient software, particularly neural networks, by providing a standardized and accessible approach that incorporates energy consumption as a critical criterion alongside accuracy in network selection.

With the increasing importance of energy efficiency in the digital era, Phoeni6 represents a significant advancement in promoting sustainable and responsible practices in the development and deployment of neural networks.

\subsection{Future Work and Research Direction}
\label{sec-future-work}
As Phoeni6 evolves, its scalability will be a key focus for future development. Efforts will be directed toward optimizing the framework for large-scale neural networks and diverse hardware configurations, ranging from edge devices to multi-GPU systems. This scalability will enable more robust energy evaluations across various platforms, ensuring the framework remains versatile and applicable to real-world scenarios. Future iterations of Phoeni6 will also incorporate additional evaluation metrics, such as Power Usage Effectiveness (PUE), carbon footprint, and water usage, allowing for a more comprehensive analysis of the environmental impact of neural networks. These enhancements will require users to provide detailed information about their execution environments, including energy sources and data center configurations.

Additionally, plans include expanding the flexibility of Phoeni6 by enabling better integration with emerging technologies, such as Internet of Things (IoT) devices, which are increasingly relevant for energy efficiency studies. Distributed computing strategies will also be explored to handle the growing complexity of neural networks and their energy profiles. This will further enhance the performance and scalability of the framework.

Another critical direction involves improving the user experience and automating more aspects of the evaluation process. By refining the orchestration of containers and expanding compatibility with diverse hardware drivers, Phoeni6 will aim to minimize manual intervention while ensuring reliable and reproducible results. These improvements are aligned with the overarching goal of supporting the adoption of sustainable and energy-efficient neural network practices in both academic and industrial settings.

By addressing these \textit{future directions} and engaging in the \textit{concrete future work} described, Phoeni6 has the potential to become a foundational tool for evaluating energy consumption in neural networks, contributing to more sustainable and responsible AI practices.

All tools that compose Phoeni6 are open-source and available at https://gitlab.com/lappsufrn/phoeni6.

\bibliographystyle{unsrtnat}
\bibliography{main}




\end{document}